\documentclass{article}

\usepackage{microtype}
\usepackage{graphicx}
\usepackage{amsmath}
\usepackage{amssymb}
\usepackage{amsthm}
\usepackage[colorlinks=true,urlcolor=blue,anchorcolor=blue,citecolor=blue,filecolor=blue,linkcolor=blue,menucolor=blue,linktocpage=true]{hyperref}
\usepackage{graphicx}
\usepackage[export]{adjustbox}
\usepackage{subfig}
\usepackage{float}
\usepackage{natbib}
\pdfoutput=1
\usepackage{soul}

\usepackage[percent]{overpic}

\newif\ifarxiv
\arxivtrue

\newcommand{\dho}{\partial}

\newcommand{\bR}{\ensuremath{\mathbb{R}}}

\newcommand{\cN}{\ensuremath{\mathcal{N}}}

\newcommand{\cO}{\ensuremath{\mathcal{O}}}

\newcommand{\etamax}{\eta_{\rm max}}

\newcommand{\pd}[2]{\frac{\partial #1}{\partial #2}}

\newcommand{\lmax}{\lambda}

\newcommand{\Dfmax}{\tilde{f}^{\rm max}}
\newcommand{\emax}{e^{\rm max}}

\ifarxiv
\usepackage[accepted]{icml2020_arxiv}
\else
\usepackage{icml2020}
\fi

\icmltitlerunning{The large learning rate phase of deep learning}
\begin{document}

\ifarxiv
\onecolumn
\fi

\icmltitle{The large learning rate phase of deep learning: \\ the catapult mechanism}
\begin{icmlauthorlist}
\icmlauthor{Aitor Lewkowycz}{google}
\icmlauthor{Yasaman Bahri}{brain}
\icmlauthor{Ethan Dyer}{google}
\icmlauthor{Jascha Sohl-Dickstein}{brain}
\icmlauthor{Guy Gur-Ari}{google}
\end{icmlauthorlist}
\icmlcorrespondingauthor{Guy Gur-Ari}{guyga@google.com}
\icmlaffiliation{google}{Google}
\icmlaffiliation{brain}{Google Brain}
\vspace*{3mm}
\printAffiliationsAndNotice{}

\begin{abstract}
The choice of initial learning rate can have a profound effect on the performance of deep networks. 
We present a class of neural networks with solvable training dynamics, and confirm their predictions empirically in practical deep learning settings.
The networks exhibit sharply distinct behaviors at small and large learning rates. The two regimes are separated by a phase transition.
In the small learning rate phase, training can be understood using the existing theory of infinitely wide neural networks. At large learning rates the model captures qualitatively distinct phenomena, including the convergence of gradient descent dynamics to flatter minima. One key prediction of our model is a narrow range of large, stable learning rates. We find good agreement between our model's predictions and training dynamics in realistic deep learning settings. Furthermore, we find that the optimal performance in such settings is often found in the large learning rate phase. We believe our results shed light on characteristics of models trained at different learning rates. In particular, they fill a gap between existing wide neural network theory, and the nonlinear, large learning rate, training dynamics relevant to practice. 
\end{abstract}

\section{Introduction}

Deep learning has shown remarkable success across a variety of machine learning tasks.
At the same time, our theoretical understanding of deep learning methods remains limited.
In particular, the interplay between training dynamics, properties of the learned network, and generalization remains a largely open problem.

\begin{figure}[ht!]
\ifarxiv
\centering
  \begin{overpic}[width=3in,trim=0 158 0 0,clip]{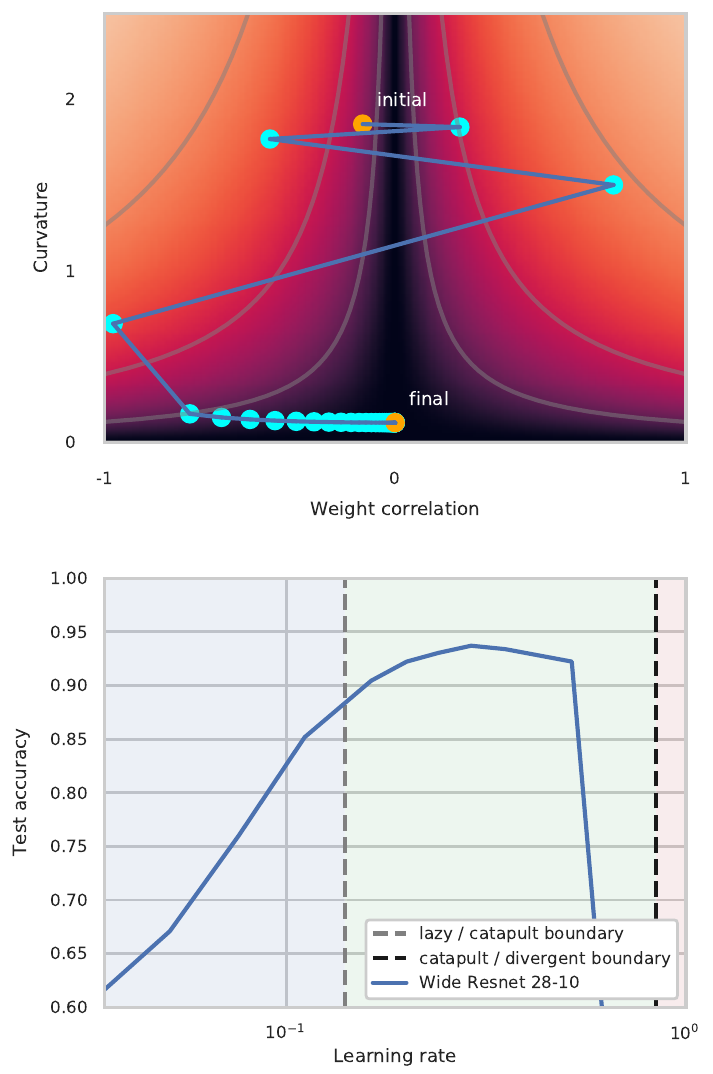}
 \put (-1,2) {\small(a)}
\end{overpic}
\qquad
  \begin{overpic}[width=3in,trim=0 -5 0 158,clip]{figures/fig1.pdf}
 \put (-1,2) {\small(b)}
\end{overpic}
\else
\centering
  \begin{overpic}[width=3in]{figures/fig1.pdf}
 \put (-1,53) {\small(a)}
 \put (-1,2) {\small(b)}
\end{overpic}
\fi
  \caption{A summary of our main results. (a) A visualization of gradient descent dynamics derived in our theoretical setup. A 2D slice of parameter space is shown, where lighter color indicates higher loss and dots represents points visited during optimization.
  Initially, the loss grows rapidly while local curvature decreases.
  Once curvature is sufficiently low, gradient descent converges to a flat minimum.
  We call this the {\em catapult effect}.
  See Figures~\ref{fig:modelfig} and \ref{fig:viz} for more details.
  (b) Confirmation of our theoretical predictions in a practical deep learning setting.
  Line shows the test accuracy of a Wide ResNet trained on CIFAR-10 
  as a function of learning rate, each trained for a fixed number of steps.
  Dashed lines show our predictions for the boundaries of the large learning rate regime (the {\em catapult phase}), where we expect optimal performance to occur.
  Maximal performance is achieved between the dashed lines, confirming our predictions.
  See Section~\ref{sec:empirical} for details.
  }
  \label{fig:marquee}
\end{figure}
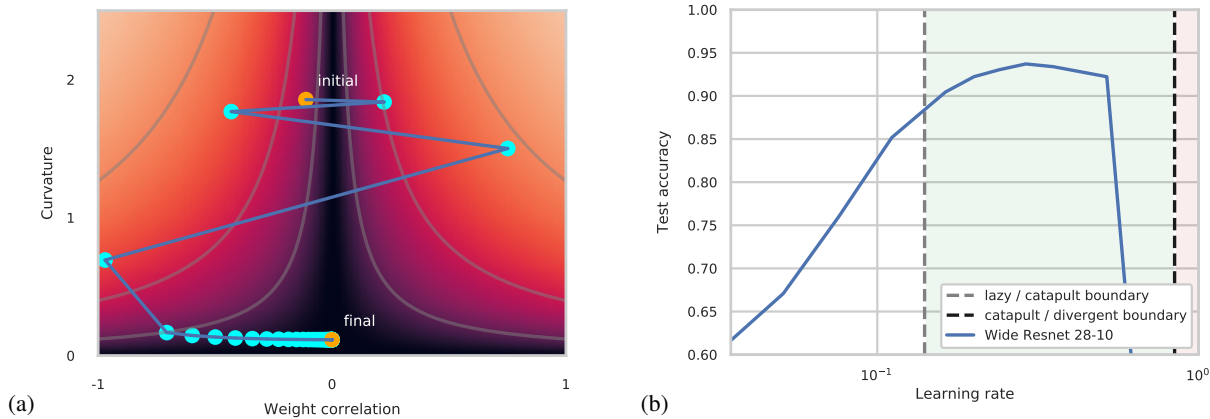

In this work we take a step toward addressing these questions.
We present a dynamical mechanism that allows deep networks trained using SGD to find flat minima and achieve superior performance.
Our theoretical predictions agree well with empirical results in a variety of deep learning settings.
In many cases we are able to predict the regime of learning rates where optimal performance is achieved.
Figure~\ref{fig:marquee} summarizes our main results.
This work builds on several existing results, which we now review.

\subsection{Large learning rate SGD improves generalization}

SGD training with large initial learning rates often leads to improved performance over training with small initial learning rates (see \citet{largelr-ma,leclerc2020regimes,xie2020diffusion,frankle2020early,jastrzebski2020break} for recent discussions).
It has been suggested that one of the mechanisms underlying the benefit of large learning rates
is that noise from stochastic gradient descent leads to flat minima, and that flat minima generalize better than sharp minima \citep{hochreiter1997, Keskar, smith2018, fantastic2020,parketal} (though see \citet{dinh2017sharp} for discussion of some caveats). 
According to this suggestion, training with a large learning rate (or with a small batch size) can improve performance because it leads to more 
stochasticity
during training \citep{mandt2017stochastic,2017arXiv171100489S,smith2018,smith2018stochastic}.

We 
will 
develop
a connection between large learning rate and flatness of minima in models trained via SGD. 
Unlike the relationship explored in most previous work though, 
this connection
is not driven by SGD noise, but arises solely as a result of training with a large initial learning rate,
and holds even for full batch gradient descent.

\subsection{The existing theory of infinite width networks is insufficient to describe large learning rates}

A recent body of work has investigated the gradient descent dynamics of deep networks in the limit of infinite width \citep{daniely2017, NTK-paper, linearized,  Du_icml2019, zou2018, AllenZhu2019, li_neurips2018, Chizat_Lazy, Mei_MeanField, rotskoff2018, sirignano2018, woodworth2019kernel, ringel2020}. Of particular relevance is the work by \citet{NTK-paper} showing that gradient flow in the space of functions is governed by a dynamical quantity  called the Neural Tangent Kernel (NTK) which is fixed at its initial value in this limit. 
\citet{linearized} showed this result is equivalent to training
the linearization of a model around its initialization in parameter space. 
Finally, moving away from the strict limit of infinite width by working perturbatively, \citet{feynman-diagrams,nth} introduced an approach to computing the finite-width corrections to network evolution. 

Despite this progress, it seems these results are insufficient to capture the full dynamics of deep networks, as well as their superior performance, in regimes applicable to practice. 
Prior work has focused on comparisons between various infinite-width kernels associated with deep networks and their finite-width, SGD-trained counterparts \citep{FC-GP,CNN-GP,Arora-CNN}. Specific findings vary depending on precise choices for architecture and hyperparameters. 
However, dramatic performance gaps are consistently observed between non-linear CNNs and their limiting kernels,
implying that the theory is not sufficient to explain the performance of deep networks in this realistic setup. 
Furthermore, some hyperparameter settings in finite-width models have no known analogue in the infinite width limit, and it is these settings that often lead to optimal performance.

In particular, 
finite width networks are often trained with large learning rates that would cause divergence for infinite width linearized models. Further, these large learning rates cause finite width networks to converge to flat minima. For infinite width linearized models, trained with MSE loss, all minima have the same curvature, and the notion of flat minima does not apply. 
We argue that the reduction in curvature during optimization, and support for learning rates that are infeasible for infinite width linearized models, may thus partially explain performance gaps observed between linear and non-linear models. 

\subsection{Our contribution: three learning rate regimes}

In this work, we identify a dynamical mechanism which enables finite-width networks to stably access large learning rates. 
We show that this mechanism causes training to converge to flatter minima and is associated with improved generalization. We further show that this same mechanism can describe the behavior of infinite width networks, if training time is increased along with network width.

This new mechanism enables a characterization of gradient descent training in terms of three learning rate regimes, or phases: the {\em lazy phase}, the {\em catapult phase}, and the {\em divergent phase}. 
In Section \ref{sec:model} we analytically derive the behavior in these three learning rate regimes for one hidden layer linear networks with large but finite width, trained with MSE loss.
We confirm experimentally in Section \ref{sec:empirical} that these phases also apply to deep nonlinear fully- connected, convolutional, and residual architectures. 
In Section~\ref{sec large width} we study additional predictions of the analytic solution.

We now summarize all three phases, using $\eta$ to indicate the learning rate, and $\lambda_0$ to indicate the initial curvature (defined precisely in Section \ref{sec warmup}).
The phase is determined by the curvature at initialization and by the learning rate, despite the fact that the curvature may change significantly during training.
Based on the experimental evidence we expect the behavior described below to apply in typical deep learning settings, when training sufficiently wide networks using SGD.

\paragraph{Lazy phase: $\eta < 2/\lambda_0$ .}
For sufficiently small learning rate, the curvature $\lambda_t$ at training step $t$ remains constant during the initial part of training. The model behaves (loosely) as a model linearized about its initial parameters \citep{linearized}; this becomes exact in the infinite width limit, where these dynamics are sometimes called \emph{lazy training} \citep{NTK-paper,linearized, Du_icml2019, li_neurips2018, zou2018, AllenZhu2019, Chizat_Lazy, feynman-diagrams}. For a discussion of trainability and the connection to the NTK in the lazy phase see \citet{xiao2019disentangling}. 

\newcommand{\carch}{c_{\mathrm{act.}}}

\paragraph{Catapult phase: $2/\lambda_0 < \eta < \etamax$ .} 

In this phase, the curvature at initialization is too high for training to converge to a nearby point, and the linear approximation quickly breaks down.
Optimization begins with a period of exponential growth in the loss, coupled with a rapid decrease in curvature, until curvature stabilizes at a value $\lambda_{\rm final} < 2/\eta$. 
Once the curvature drops below $2/\eta$, training converges, ultimately reaching a minimum that is flatter than those found in the lazy phase.
This initial period lasts for a number of training steps that is of order $\log(n)$, where $n$ is the network width, and is therefore quite short for realistic networks (often lasting less than a single epoch). Optimal performance is often achieved when the initial learning rate is in this range.
The gradient descent dynamics in this phase are visualized in SM Figure~\ref{fig:viz} and in Figure~\ref{fig:marquee}.

The maximum learning rate is approximately given by $\etamax=\carch/\lambda_0$, where $\carch$ is an architecture-dependent constant.
Empirically, we find that this constant depends strongly on the non-linearity but only weakly on other aspects of the architecture.
For networks with ReLU non-linearity we find empirically that $\carch \approx 12$.
For the theoretical model, we show that $\carch=4$.

\paragraph{Divergent phase: $\eta > \etamax$ .} When the learning rate is above the maximum learning rate of the model, the loss diverges and the model does not train.

\section{Theoretical results}
\label{sec:model}

\newcommand{\Df}{\tilde{f}}
\newcommand{\DTheta}{\tilde{\Theta}}
\newcommand{\etacrit}{\eta_{\rm crit}}

We now present our main theoretical result, an analysis of gradient descent dynamics for a neural network with large but finite width.

Given a network function $f:\bR^d \to \bR$ with model parameters $\theta\in\bR^p$, and a training set $\{(x_\alpha,y_\alpha)\}_{\alpha=1}^m$, the MSE loss is
\begin{align}
    L = \frac{1}{2m} \sum_{\alpha=1}^m (f(x_\alpha) - y_\alpha)^2 \,. \label{eq:loss}
\end{align}

The NTK
$\Theta:\bR^d \times \bR^d \to \bR$ is defined by
\begin{align}\label{eq:NTKdef}
  \Theta(x,x') &:= \frac{1}{m} \sum_{\mu=1}^p \frac{\dho f(x)}{\dho \theta_\mu} \frac{\dho f(x')}{\dho \theta_\mu} \,.
\end{align}
We denote by $\lambda$ the maximum eigenvalue of the kernel.
In large width models, $\lambda$ provides a local measure of the loss landscape curvature that is similar to the top eigenvalue of the Hessian \citep{feynman-diagrams}.

In this section, we will consider a network with one hidden layer and linear activations, where the network function $f$ is given by
\begin{align}
  f(x) = n^{-1/2} v^T u x \,. \label{eq:f}
\end{align}
Here $n$ is the width (number of neurons in the hidden layer), $v \in \bR^{n}$ and $u \in \bR^{n\times d}$ are the model parameters (collectively denoted $\theta$), and $x\in\bR^{d}$ is the training input.
At initialization, the weights are drawn from $\cN(0,1)$.

\subsection{Warmup: a simplified model}
\label{sec warmup}
Before analyzing the dynamics of the model, we analyze a simpler setting which captures 
the most important aspects of the full solution.
Consider a dataset with 1D inputs, and with a single training sample $x=1$ with label $y=0$.
The network function evaluated on this input is then $f = n^{-1/2} v^T u$, with $u,v \in \bR^n$, and the loss is $L = f^2/2$.
The gradient descent equations at training step $t$ are
\begin{align}
  u_{t+1} &= u_{t} - \eta n^{-1/2} f_t v_{t} \,, \;
  v_{t+1} = v_{t} - \eta n^{-1/2} f_t u_{t} \,.
   \label{eq:gd_weights}
\end{align}
Next, consider the update equations in function space.
These can be written in terms of the Neural Tangent Kernel. 
For this model, the kernel evaluated on the training set is a scalar which is equal to $\lambda$, its top eigenvalue, and is given by
\begin{align}\label{eq:NTK}
  \Theta(1,1) =
  \lambda &= n^{-1} \left( \| v \|_2^2 + \| u \|_2^2 \right) \,.
\end{align}
At initialization, 
both $f^2$ and $\lambda$ scale as $n^0 = 1$ with width. 
The following update equations for $f$ and $\lambda$ at step $t$ can be derived from \eqref{eq:gd_weights}.
\begin{align}
  f_{t+1} &= 
  \left( 1 - \eta \lambda_t + \frac{\eta^{2} f_t^2}{n} \right) f_t
  \,, \label{eq:fsimpupdate_f} \\
  \lambda_{t+1} &= \lambda_{t} + 
  \frac{\eta f_t^{2}}{n} \left( \eta\lambda_t - 4 \right)
  \,. \label{eq:fsimpupdate_l} 
\end{align}
It is important to note that these are the exact update equations for this model, and that no higher-order terms were neglected.
We now analyze these dynamical equations assuming the width $n$ is large.
Two learning rates that will be important in the analysis are $\etacrit=2/\lambda_0$ and $\etamax=4/\lambda_0$.
In terms of the notation introduced above, the architecture-dependent constant that determines that maximum learning rate in this model is $\carch=4$.

\subsubsection{Lazy phase}
Taking the strict infinite width limit, equations \eqref{eq:fsimpupdate_f} and \eqref{eq:fsimpupdate_l} become
\begin{align}
f_{t+1} &= \left( 1 - \eta \lambda_t \right) f_t \,, \quad
\lambda_{t+1} = \lambda_{t} \label{eq:fsimplwupdate} \,.
\end{align}
When $\eta < \etacrit$, $\lambda$ remains constant throughout training.
This is a special case of NTK dynamics, where the kernel is constant and the network evolves as a linear model \citep{linearized}.
The function and the loss both shrink to zero because the multiplicative factor obeys $|1 - \eta \lambda_t| < 1$. This convergence happens in $\cO(n^0) = \cO(1)$ steps.

\subsubsection{Catapult phase}
When $\etacrit < \eta < \etamax$, the loss diverges in the infinite width limit.
Indeed, from \eqref{eq:fsimplwupdate} we see that the kernel is constant in the limit, while $f$ receives multiplicative updates where $|1 - \eta \lambda_t| > 1$.
This is the well known instability of gradient descent dynamics for linear models with MSE loss.
However, the underlying model is not linear in its parameters, and finite width contributions turn out to be important.
We therefore relax the infinite width limit and analyze equations (\ref{eq:fsimpupdate_f},\ref{eq:fsimpupdate_l}) for large but finite width, $n \gg 1$.

First, note that $\eta \lambda_0 - 4 < 0$ by assumption, and therefore the (additive) kernel updates are negative for all $t$.
During early training, $|f_t|$ grows (as in the infinite width limit) while $\lambda_t$ remains constant up to small $\cO(n^{-1})$ updates.
After $t \sim \log(n)$ steps, $|f_t|$ grows to order $n^{1/2}$.
At this point, the kernel updates are no longer negligible because $f_t^2 / n$ is of order $n^0$.
The kernel $\lambda_t$ receives negative, non-negligible updates while both $f_t$ and the loss continue to grow (for now, we ignore the term in \eqref{eq:fsimpupdate_f} with an explicit $1/n$ dependence).
This continues until the kernel is sufficiently small that the condition $\eta \lambda_t \lesssim 2$ is met.\footnote{The bound is not exact because of the term we neglected.}
We call this curvature-reduction effect the {\em catapult effect}.
Beyond this point, $|1 - \eta \lambda_t| < 1$ holds, $|f_t|$ shrinks, and the loss converges to a global minimum.
The $n$ dependence of the steps until optimization converges is $\log{(n)}$.

It remains to show that the term in \eqref{eq:fsimpupdate_f} with an explicit $n^{-1}$ dependence does not affect these conclusions.
Once $|f_t|$ grows to order $n^{1/2}$, this term is no longer negligible and can cause the multiplicative factor in front of $f_t$ to become smaller than 1 in absolute value, causing $|f_t|$ to start shrinking.
However, once $|f_t|$ shrinks sufficiently this term again becomes negligible.
Therefore, the loss will not converge to zero unless the curvature eventually drops below $2/\eta$.
Conversely, notice that this term cannot cause $|f_t|$ to diverge for learning rates below $\etamax$.
Indeed, if this were to happen then equation \eqref{eq:fsimpupdate_l} would drive $\lambda_t$ to negative values, leading to a contradiction.
This completes the analysis in this phase.

Let us make a few comments about the catapult phase. 

It is important for the analysis that we take a modified large width limit, in which the number of training steps grows like $\log(n)$ as $n$ becomes large. 
This is different than the large width limit commonly studied in the literature, in which the number of steps is kept fixed as the width is taken large.
When using this modified limit, the analysis above holds even in the limit.
Note as well that the catapult effect takes place over $\log(n)$ steps, and for practical networks will occur within the first 100 steps or so of training.

In the catapult phase, the kernel at the end of training is smaller by an order $n^0$ amount
compared with its value at initialization.
The kernel provides a local measure of the loss curvature. 
Therefore, the minima that SGD finds in the catapult phase are flatter than those it finds in the lazy phase.
Contrast this situation, in which the kernel receives non-negligible updates, with the conclusions of \citet{NTK-paper} where the kernel is constant throughout training.
The difference is due to the large learning rate, which leads to a breakdown of the linearized approximation even at large width.

Figure~\ref{fig:modelfig} illustrates the dynamics in the catapult phase.
For learning rates $\etacrit < \eta < \etamax$ we observe the catapult effect: the loss goes up before converging to zero.
The curvature exhibits the expected sharp transitions as a function of the learning rate: it is constant in the lazy phase, decreases in the catapult phase, and diverges for $\eta > \etamax$.
\ifarxiv
\begin{figure}[ht!]
  \subfloat[]{\includegraphics[width=0.33 \textwidth]{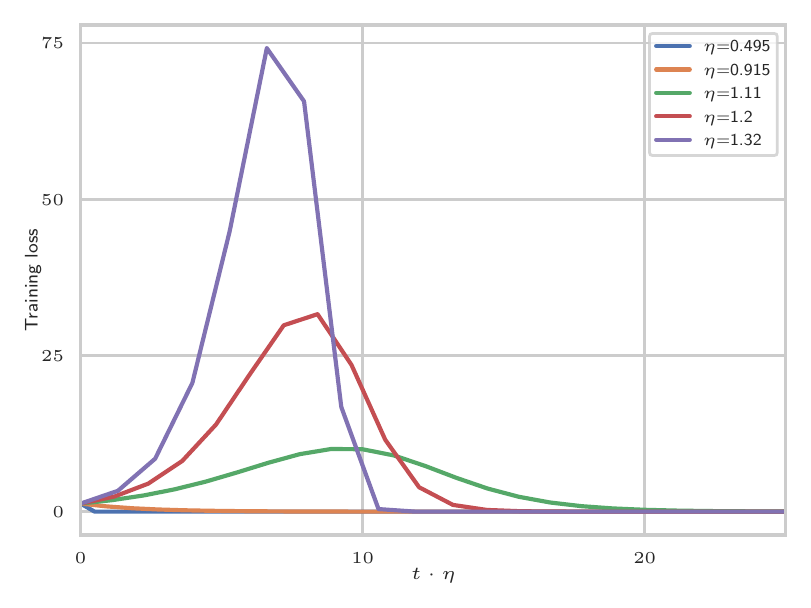}}
  \subfloat[]{\includegraphics[width=0.33\textwidth]{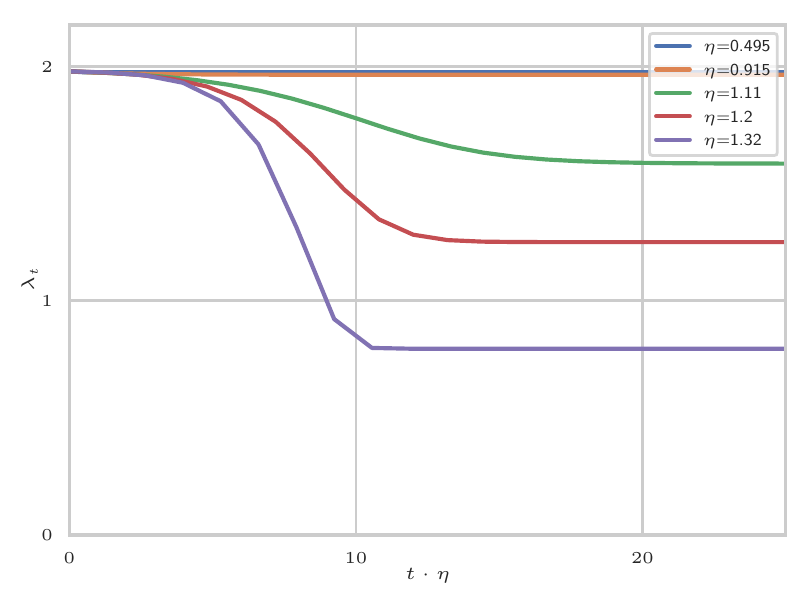}}
  \subfloat[]{\includegraphics[width=0.33\textwidth]{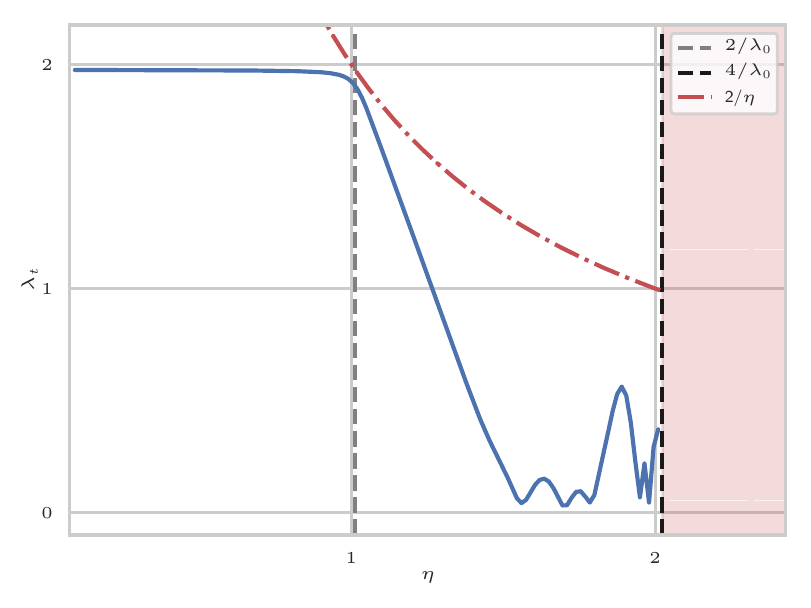}}
  \caption{ Empirical results for the gradient descent dynamics of the warmup model with $n=10^3$, for which $\etacrit \approx 1$. (a) Training loss for different learning rates.
    (b) Maximum NTK eigenvalue as a function of time. For $\eta > 1$, $\lambda_t$ decreases rapidly to a fixed value.
    (c) Maximum NTK eigenvalue at $t = 25/\eta$. The shaded area indicates learning rates for which training diverges empirically. 
    The results are presented as a function of $t \cdot \eta$ (rather than $t$) for convenience.
  }
  \label{fig:modelfig}
\end{figure}
\else
\begin{figure}[ht!]
  \subfloat[]{\includegraphics{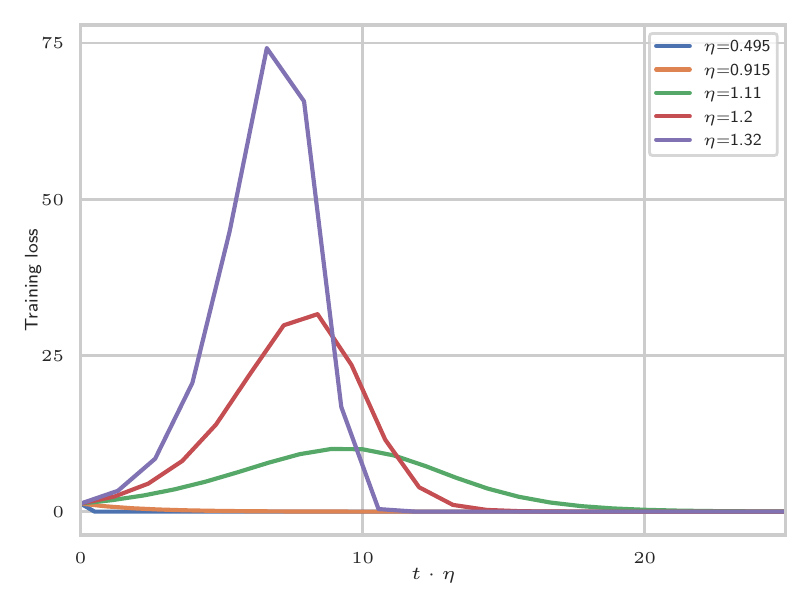}}
  \newline
  \subfloat[]{\includegraphics{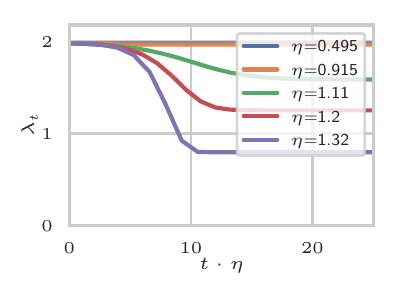}}
  \subfloat[]{\includegraphics{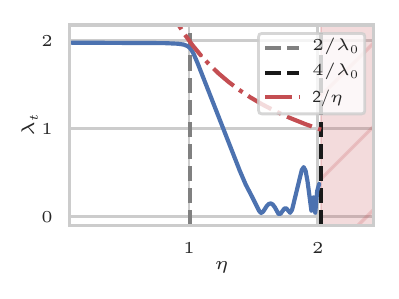}}
  \caption{ Empirical results for the gradient descent dynamics of the warmup model with $n=10^3$, for which $\etacrit \approx 1$. (a) Training loss for different learning rates.
    (b) Maximum NTK eigenvalue as a function of time. For $\eta > 1$, $\lambda_t$ decreases rapidly to a fixed value.
    (c) Maximum NTK eigenvalue at $t = 25/\eta$. The shaded area indicates learning rates for which training diverges empirically. 
    The results are presented as a function of $t \cdot \eta$ (rather than $t$) for convenience.
  }
  \label{fig:modelfig}
\end{figure}
\fi
\subsubsection{Divergent phase}
Completing the analysis of this model, when $\eta > \etamax$ the loss diverges because the kernel receives positive updates, accelerating the rate of growth of the function.
Therefore, $\etamax=4/\lambda_0$ is the maximum learning rate of the model.

\subsection{Full model}
\label{sec:full_model}

We now turn to analyzing the model presented at the beginning of this section, with $d$-dimensional inputs and $m$ training samples with general labels.
The full analysis is presented in SM Section~\ref{sec:sm:fullmodel}; here we summarize the argument.
The conclusions are essentially the same as those of the warmup model.

We introduce the notation $f_\alpha := f(x_\alpha)$ for the function evaluated on a training sample, $\Df_\alpha := f_\alpha - y_\alpha$ for the error, and $\Theta_{\alpha\beta} := \Theta(x_\alpha, x_\beta)$ for the kernel elements.
We will treat $f,\Df$ evaluated on the training set as vectors in $\bR^m$, whose elements are $f_\alpha,\Df_\alpha$.
Consider the following update equation for the error, which can be derived from the update equations for the parameters. 
Note that this is the exact update equation for this model; no higher-order terms were neglected.
\begin{align}
  \Df_\alpha^{t+1} &= \sum_\beta (\delta_{\alpha\beta} - \eta \Theta_{\alpha\beta}) \Df_\beta + \frac{\eta^2}{nm} (x_\alpha^T \zeta)
  (f^T \Df) \,. \label{eq:fupfull}
\end{align}
Here, $\zeta := \sum_\alpha \Df_\alpha x_\alpha / m \in \bR^d$, and all variables are implicitly evaluated at step $t$ unless specified otherwise.

We again take the modified large width limit $n \to \infty$, allowing the number of steps to scale logarithmically in the width.
At initialization, $f_\alpha$, $\Df_\alpha$, and $\Theta_{\alpha\beta}$ are all of order $n^0$.
We now analyze the gradient descent dynamics as a function of the learning rate.

The maximum eigenvalue of the kernel at step $t$ is $\lambda_t$.
When $\eta < \etacrit$, the norm $\| \Df^t \|_2$ shrinks to zero in $\cO(n^0)$ time while the kernel receives $\cO(n^{-1})$ corrections.
Therefore, in the limit the kernel remains constant until convergence.
This is a special case of the NTK result \citep{NTK-paper}, and the model evolves as a linear model.

Next, suppose that $\etacrit < \eta < \etamax$.
Early during training $\| \Df \|_2$ grows, with the fastest growth taking place along the direction of the top kernel eigenvector, $\emax_t \in \bR^m$.
During this part of training the kernel receives $\cO(n^{-1})$ updates, and so $\emax_t$ does not change much.
As a result, $\Df_t$ becomes aligned with $\emax_t$.
In addition, $f_t$ becomes close to $\Df_t$ because $f_t$ grows while the label is constant. 
We therefore consider the following approximate update equations for $\Dfmax := \sum_\alpha \Df_\alpha \emax_\alpha$ and for the maximum eigenvalue $\lambda$, which can be approximated by $\Df^T \Theta \Df / \| \Df\|_2^2$.
\begin{align}
\Dfmax_{t+1}&\approx \left( 1-\eta \lmax_t 
\right)  \Dfmax_{t} + \cO(n^{-1}) \label{eq:fupdate}\,,\\
\lmax_{t+1} &\approx \lmax_t + \frac{\eta \|\zeta\|_2^2}{n}(\eta \lmax_t - 4) \,.\label{eq:Thetaupdate}
\end{align}
We note in passing the similarity between these equations and \eqref{eq:fsimpupdate_f}, \eqref{eq:fsimpupdate_l}.
We see that once $\Dfmax$ and $\zeta$ become of order $n^{1/2}$, $\lambda_t$ receives non-negligible negative corrections of order $n^0$.
This evolution continues until $\lambda_t \lesssim 2/\eta$, after which the error converges to zero.
Finally, if $\eta > \etamax$, the error grows while $\lambda_t$ receives positive updates, and the loss diverges. 
This concludes the discussion of the theoretical model; 
further details can be found in Section~\ref{sec large width} and in SM Section~\ref{sec:sm:fullmodel}.

\section{Experimental results}
\label{sec:empirical}

In this section we test the extent to which the behavior of our theoretical model describes the dynamics of deep networks in practical settings.
The theoretical results of Section~\ref{sec:model}, describing distinct learning rate phases, are not guaranteed to hold beyond the model analyzed there.
We treat these results as predictions to be tested empirically, including the values $\etacrit$ and $\etamax$ of the learning rates that separate the three phases.

In a variety of deep learning settings, we find clear evidence of the different phases predicted by the model.
The experiments all use MSE loss, sufficiently wide networks, 
and SGD\footnote{While our theoretical framework focused on (full-batch) gradient descent, we expect these the phases to happen at similar points for SGD as long as evolution is not noise dominated, in which case we expect all phases to be shifted towards smaller learning rates.}.
Parameters such as network architecture, choice of non-linearity, weight parameterization, and regularization, do not significantly affect this 
conclusion. 

In terms of the learning rates that determine the location of the transitions, the only modification needed to obtain good agreement with experiment is to replace the theoretical maximum learning rate, $4/\lambda_0$, with a 1-parameter function $\etamax=\carch/\lambda_0$, where $\carch$ is an architecture-dependent constant.
We find that $\carch \approx 12$ for all network that use ReLU non-linearity, and it seems this parameter depends only weakly on other details of the architecture.
We find the level of agreement with the experiments surprising, given that our theoretical model involves a shallow network without non-linearities.

Building on the observed correlation between lower curvature and generalization performance \citep{Keskar,fantastic2020}, we conjecture that optimal performance occurs in the large learning rate (catapult) phase, where the loss converges to a flatter minimum. 
For a fixed amount of computational budget, we find that this conjecture holds in all cases we tried.
Even when comparing different learning rates trained for a fixed amount of \emph{physical time} $t_{\rm phys} = t \cdot \eta$, we find that performance of models trained in the catapult phase either matches or exceeds that of models trained in the lazy phase.

\subsection{Early time curvature dynamics}

Our theoretical model makes detailed predictions for the gradient descent evolution of $\lambda$, the top eigenvalue of the NTK. 
Here we test these predictions against empirical results in a variety of deep learning models (see the Supplement for additional experimental results).

Figure~\ref{fig:FC} shows $\lambda$ during the early part of training for two deep learning settings. 
The results are compared against the theoretical predictions of a phase transition at $\etacrit=2/\lambda_0$, and a maximum learning rate of $4/\lambda_0$.
Here $\lambda_0$ is the top eigenvalue of the empirical NTK at initialization.

For learning rates $\eta < \etacrit$, we find that $\lambda$ is independent of the learning rate and constant throughout training, as expected in the lazy phase.
For $\etacrit < \eta < 4/\lambda_0$ we find that $\lambda$ decreases during training to below $2/\eta$, matching the predicted behavior in the catapult phase (note that in the Wide ResNet example, $\lambda$ initially increases before reaching its stable value).

The large learning rate behavior predicted by the model appears to persist up to the maximum learning rate, which is larger in these experiments than in the theoretical model.
In these and other experiments involving ReLU networks, we find that $\etamax \approx 12/\lambda_0$ is a good predictor of the maximum learning rate (in the SM \ref{sec:tanh} we discuss other nonlinearities). 
We conjecture that this is the typical maximum learning rate of networks with ReLU non-linearities.

Figure \ref{fig:FC} also shows the loss initially increasing before converging in the catapult phase, confirming another prediction of the model.
This transient behavior is very short, taking less than 10 steps to complete.
\ifarxiv

\begin{figure}[ht!]
  \centering
 \hspace*{-0.8cm} \subfloat[]{\includegraphics[width=0.35\textwidth]{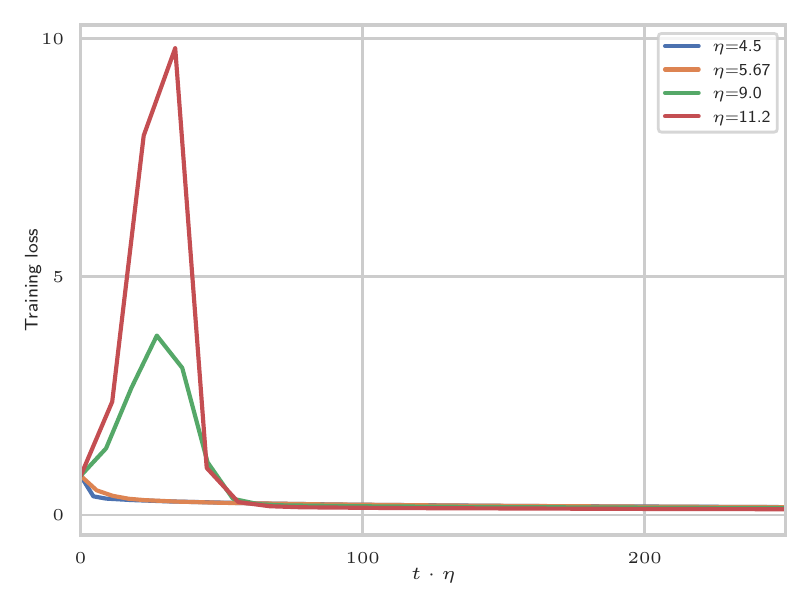}
    \label{fig:FCa}}
  \subfloat[]{
    \includegraphics[width=0.35\textwidth]
    {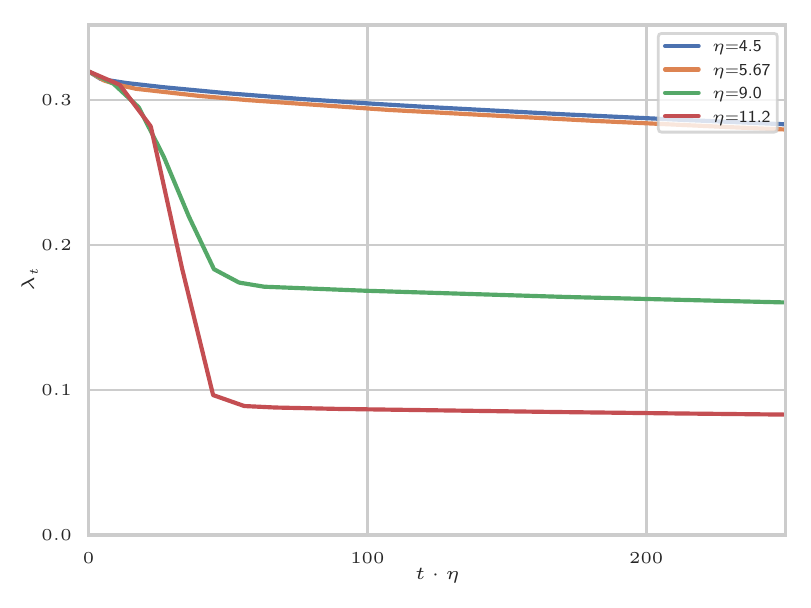}
    \label{fig:FCc}
  }
    \subfloat[]{
    \includegraphics[width=0.35\textwidth] 
    {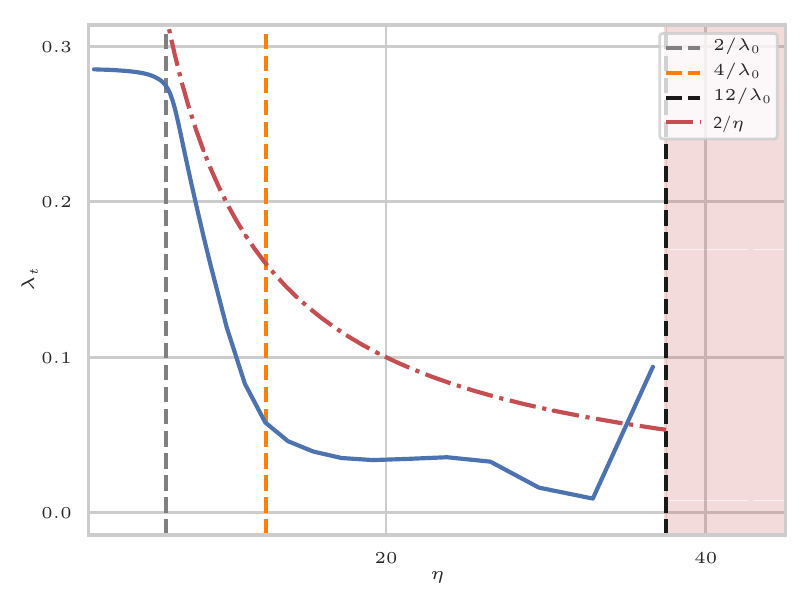}
    \label{fig:FCe}
  }
  \newline

\hspace*{-0.8cm} \subfloat[]{
  \includegraphics[width=0.35\textwidth]{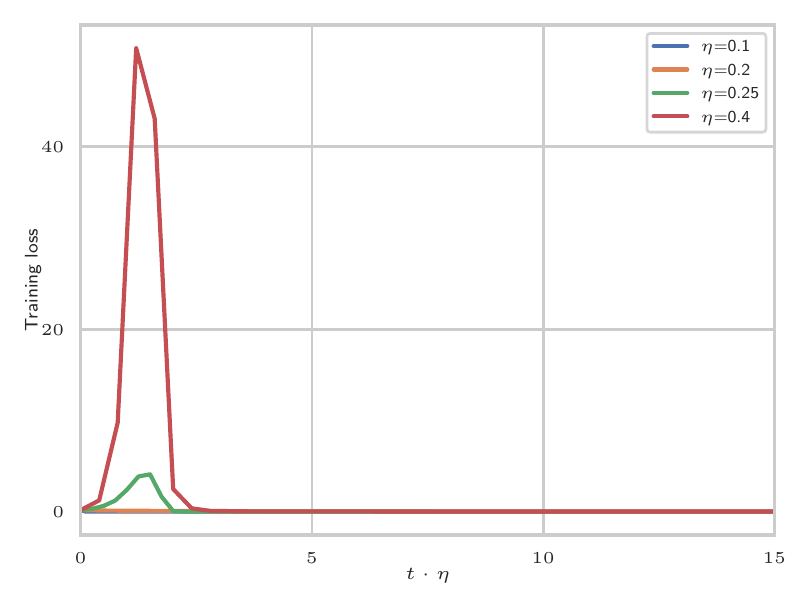}
  \label{fig:FCb}
} 
    \subfloat[]{
    \includegraphics[width=0.35\textwidth]
    {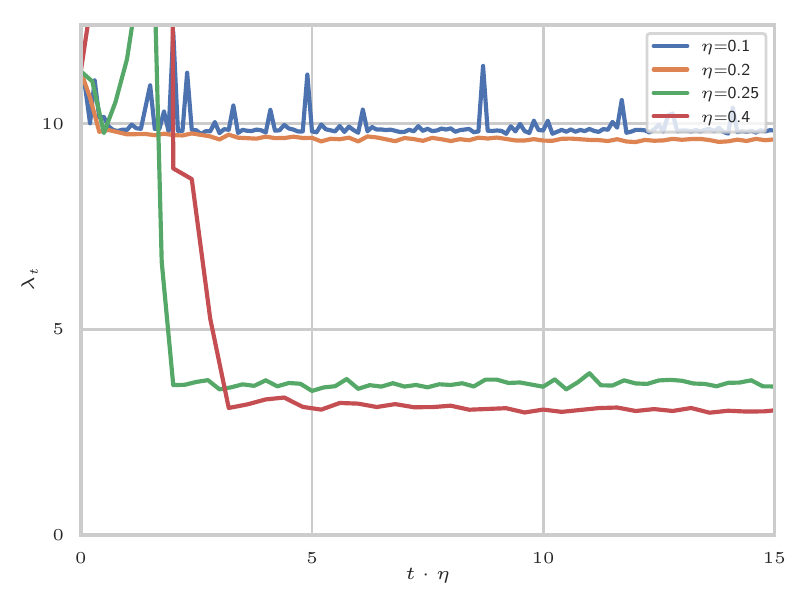}
    \label{fig:FCd}
  }
    \subfloat[]{
    \includegraphics[width=0.35 \textwidth]
    {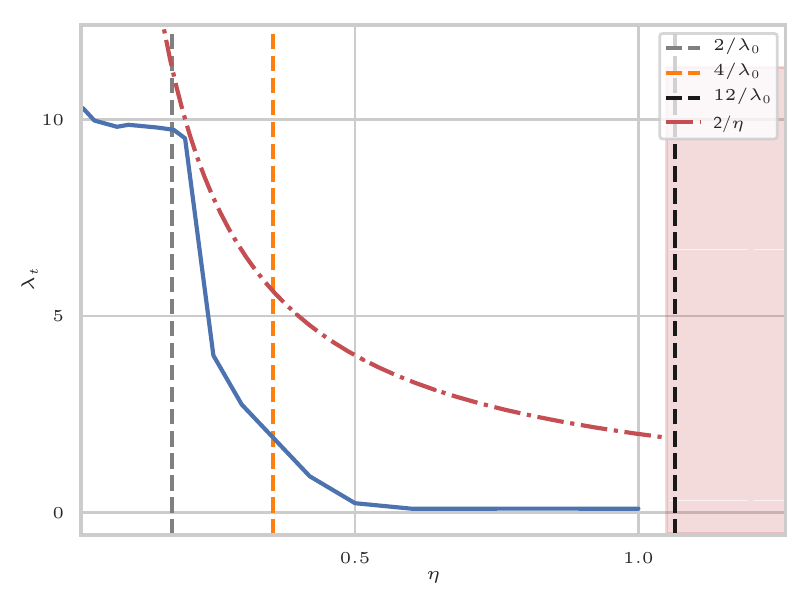}
    \label{fig:FCf}
  }
  
  \caption{Early time dynamics.
  (a,b,c) A 3 hidden layer fully-connected network with ReLU non-linearity trained on MNIST ($\eta_{\rm crit}=6.25$). (d,e,f) Wide ResNet 28-10 trained on CIFAR-10 ($\eta_{\rm crit}=0.18$). Both networks are trained with vanilla SGD; for more experimental details see SM Section~\ref{sec:expdetails}.
     (a,d) Early time dynamics of the training loss for learning rates in the linear and catapult phases.
    (b,e) Early time dynamics of the curvature for learning rates in the linear and catapult phase.
    (c,f) $\lambda_t$ measured at $t\cdot\eta=250$ (for FC) and $t\cdot \eta=30$ (for WRN), as a function of learning rate, compared with theoretical predictions for the locations of phase transitions.
    Training diverges for learning rates in the shaded region.
}\label{fig:FC}
\end{figure}
\else
\begin{figure}[ht!]
  \centering
  \subfloat[]{\includegraphics{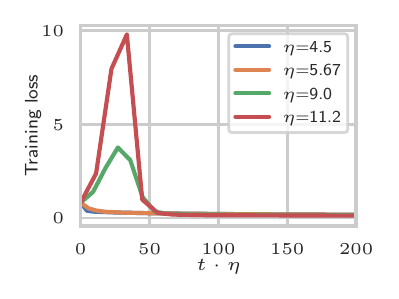}
    \label{fig:FCa}
}
\subfloat[]{
  \includegraphics{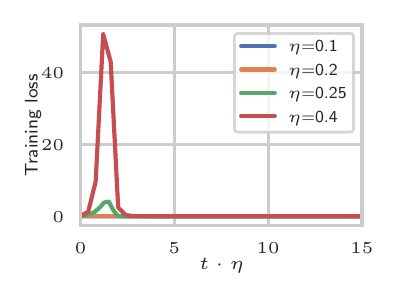}
  \label{fig:FCb}
} 
\newline \vspace{-2mm}
  \subfloat[]{
    \includegraphics
    {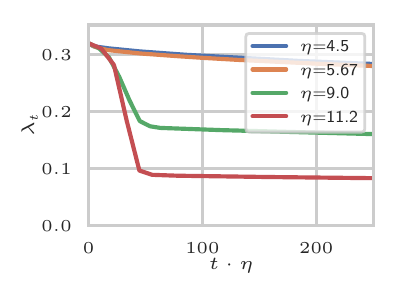}
    \label{fig:FCc}
  }
    \subfloat[]{
    \includegraphics 
    {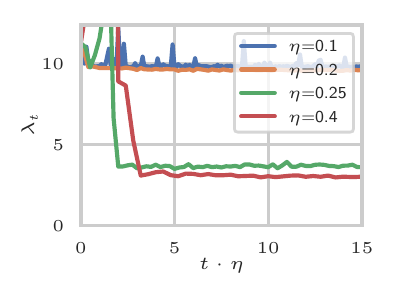}
    \label{fig:FCd}
  }
  \newline \vspace{-2mm}
  \subfloat[]{
    \includegraphics 
    {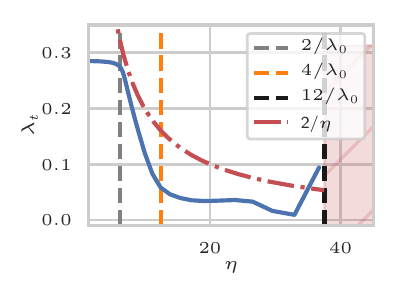}
    \label{fig:FCe}
  }
    \subfloat[]{
    \includegraphics
    {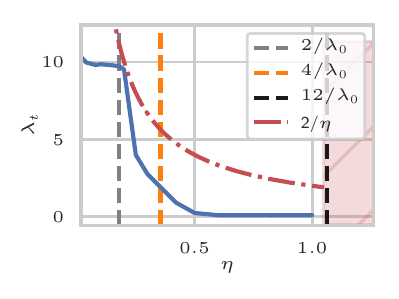}
    \label{fig:FCf}
  }
  
  \caption{Early time dynamics.
  (a,c,e) A 3 hidden layer fully-connected network with ReLU non-linearity trained on MNIST ($\eta_{\rm crit}=6.25$). (b,d,f) Wide ResNet 28-10 trained on CIFAR-10 ($\eta_{\rm crit}=0.18$). Both networks are trained with vanilla SGD; for more experimental details see SM Section~\ref{sec:expdetails}.
     (a,b) Early time dynamics of the training loss for learning rates in the linear and catapult phases.
    (c,d) Early time dynamics of the curvature for learning rates in the linear and catapult phase.
    (e,f) $\lambda_t$ measured at $t\cdot\eta=250$ (for FC) and $t\cdot \eta=30$ (for WRN), as a function of learning rate, compared with theoretical predictions for the locations of phase transitions.
    Training diverges for learning rates in the shaded region.
}\label{fig:FC}
  
\end{figure}
\fi

\subsection{Generalization performance}

We now consider the performance of trained models in the different phases discussed in this work.
\citet{Keskar} observed a correlation between the flatness of a minimum found by SGD and the generalization performance (see \citet{fantastic2020} for additional empirical confirmation of this correlation).
In this work, we showed that the minima SGD finds are flatter in the catapult phase, as measured by the top kernel eigenvalue.
Our measure of flatness differs from that of \citet{Keskar}, but we expect that these measures are correlated.

We therefore conjecture that optimal performance is often obtained for learning rates above $\etacrit$ and below the maximum learning rate.

In this section we test this conjecture empirically.
We find that performance in the large learning rate range always matches or exceeds the performance when $\eta < \etacrit$. 
For a fixed compute budget, we find that the best performance is always found in the catapult phase.

Figure~\ref{fig:fcperfgap} shows the accuracy as a function of the learning rate for a fully-connected ReLU network trained on a subset of MNIST.
We find that the optimal performance is achieved above $\etacrit$ and close to $\etamax=12/\lambda_0$, the expected maximum learning rate.

\ifarxiv
\begin{figure}[ht]
\centering
\subfloat{
  \includegraphics{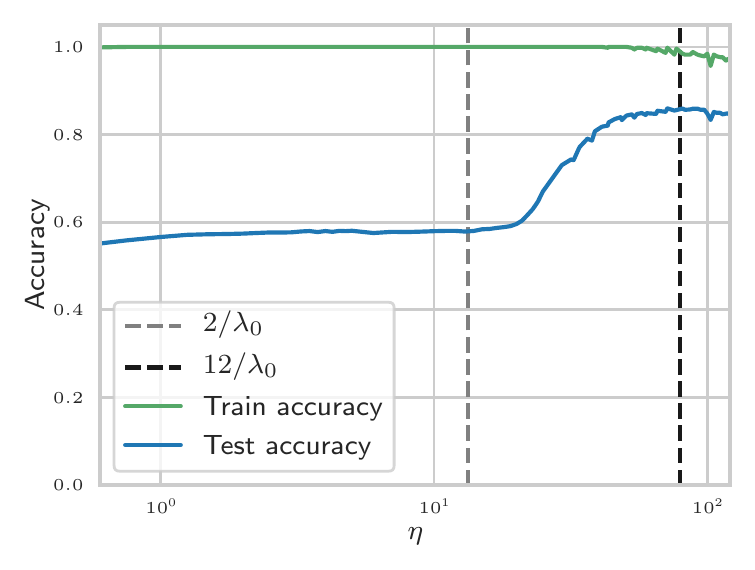}
}
\caption{Final accuracy versus learning rate 
for a fully-connected 1 hidden layer ReLU network, trained on 512 samples of MNIST with full-batch gradient descent until training accuracy reaches 1 or 700k physical steps (see SM Section~\ref{sec:expdetails} for details).
We used a subset of samples to accentuate the performance difference between phases.
The optimal performance is obtained when the learning rate is above $\etacrit$, and close to $\etamax$.
}
\label{fig:fcperfgap}
\end{figure}
\else
\begin{figure}[ht]
\centering
\subfloat{
  \includegraphics 
  {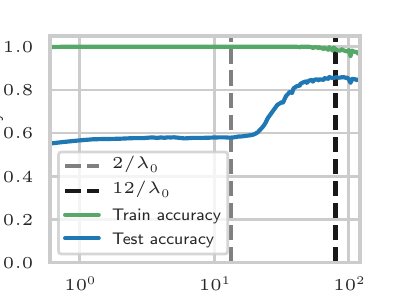}
}
\caption{Final accuracy versus learning rate 
for a fully-connected 1 hidden layer ReLU network, trained on 512 samples of MNIST with full-batch gradient descent until training accuracy reaches 1 or 700k physical steps (see SM Section~\ref{sec:expdetails} for details).
We used a subset of samples to accentuate the performance difference between phases.
The optimal performance is obtained when the learning rate is above $\etacrit$, and close to $\etamax$.
}
\label{fig:fcperfgap}
\end{figure}
\fi

\begin{figure*}[ht!]
\ifarxiv
\centering
\subfloat[]{
  \includegraphics{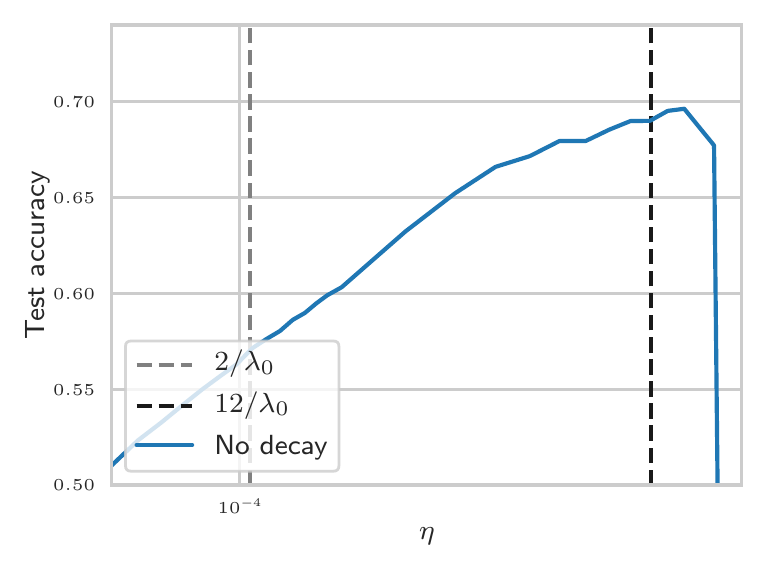}
  \label{fig:cnn_fixed}
}
\subfloat[]{
  \includegraphics{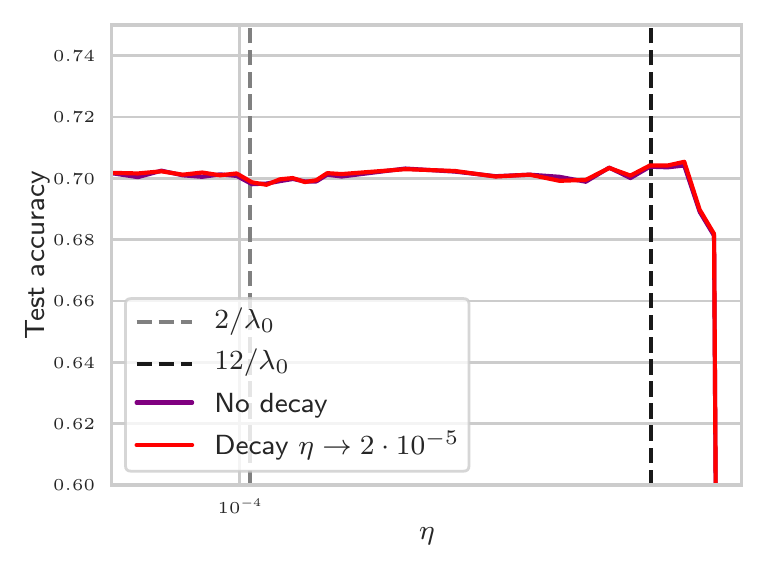}
  \label{fig:cnn_phys}
}
\newline
\centering
\subfloat[]{
  \includegraphics{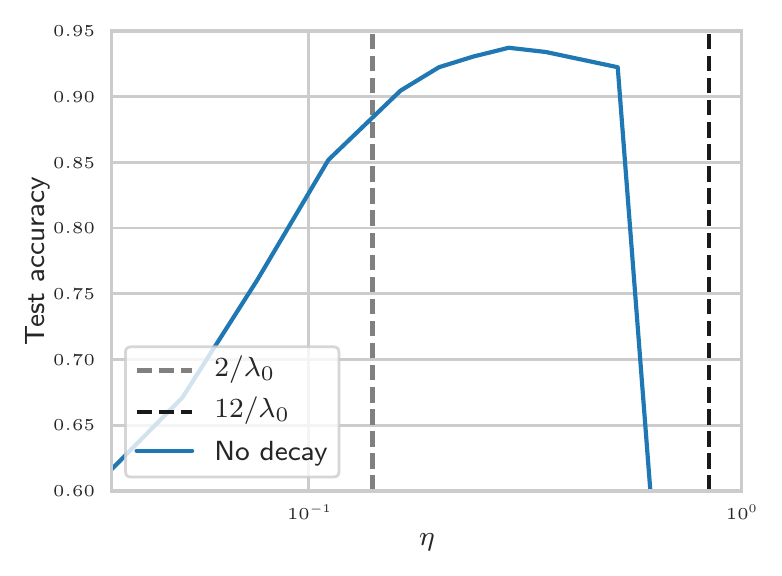}
\label{fig:wrnL2_fixed}
}
\subfloat[]{
  \includegraphics{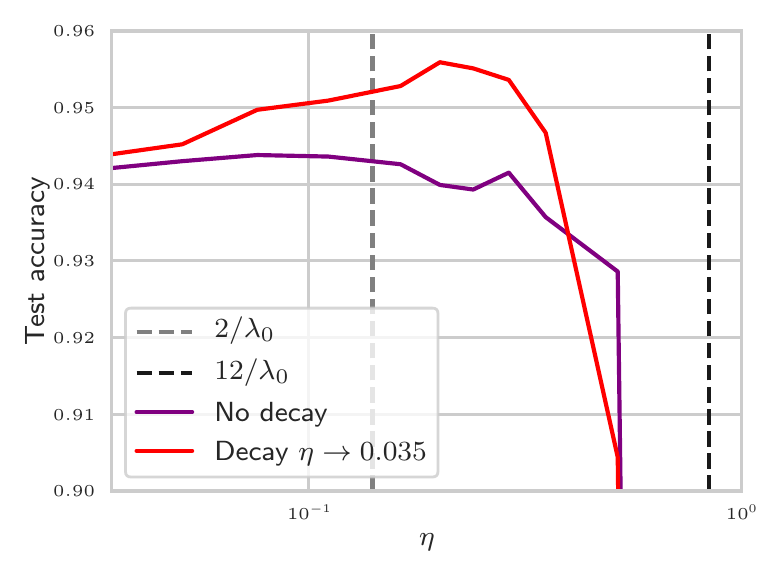}
\label{fig:wrnL2_phys}
}
\vspace{-2mm}
\else
\hspace*{-2.8mm}
\subfloat[]{
  \includegraphics{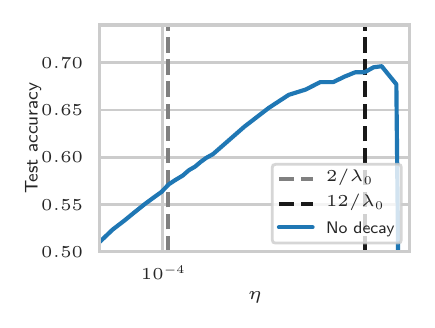}
  \label{fig:cnn_fixed}
}
\hspace*{-2.8mm}\subfloat[]{
  \includegraphics{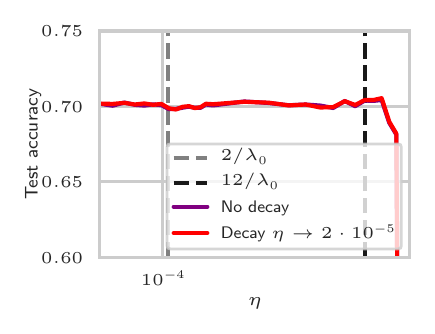}
  \label{fig:cnn_phys}
}
\hspace*{-2.8mm}\subfloat[]{
  \includegraphics{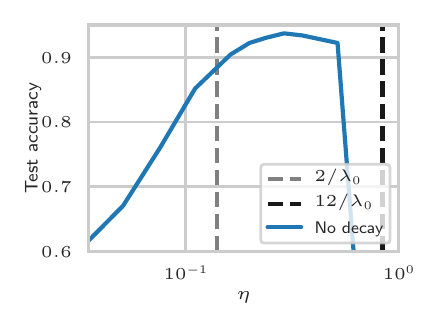}
\label{fig:wrnL2_fixed}
}
\hspace*{-2.8mm}\subfloat[]{
  \includegraphics{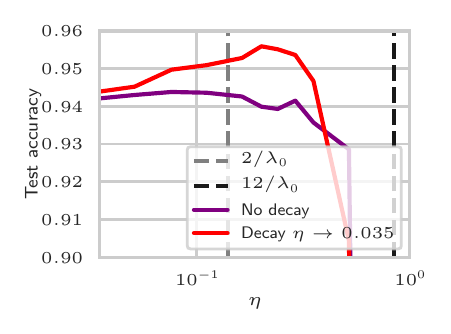}
\label{fig:wrnL2_phys}
}
\vspace{-2mm}
\fi
\caption{ Test accuracy vs learning rate for (a,b) a CNN trained on CIFAR-10 using SGD with batch size 256 and $L_{2}$ regularization  ($\etacrit\approx 10^{-4}$) and (c,d) WRN28-10 trained on CIFAR-10 using SGD with batch size 1024, $L_2$ regularization, and data augmentation  ($\etacrit\approx 0.14$); see SM \ref{sec:expdetails} for details. (a,c) have a fixed compute budget: (a) 437k steps and (b)  12k steps.  (b,d) have been evolved for a fixed amount of physical time: (b) was evolved for 475$/\eta$ steps (purple) and evolved for 50k more steps at learning rate $2 \cdot 10^{-5}$ (red) and (d) was evolved for $3360/\eta$ steps with learning rate $\eta$ (purple) and then evolved for 4800 more steps at learning rate $0.035$ (red).
In all cases, optimal performance is achieved above $\etacrit$ and close to the expected maximum learning rate, in agreement with our predictions.
}
\label{fig:cifar10perf}

\end{figure*}

Next, Figure~\ref{fig:cifar10perf} shows the performance of a convolutional network and a Wide ResNet (WRN) trained on CIFAR-10.
The experimental setup, which we now describe, was chosen to ensure a fair comparison of the performance across different learning rates.
The network is trained with different initial learning rates, followed by a decay at a fixed physical time $t\cdot\eta$ to the same final learning rate.
This schedule is introduced in order to ensure that all experiments have the same level of SGD noise toward the end of training.

We present results using two different stopping conditions.
In Figure~\ref{fig:cnn_fixed}, \ref{fig:wrnL2_fixed}, all models were trained for a fixed number of training steps.
We find a significant performance gap between small and large learning rates, with the optimal learning rate above $\etacrit$ and close to $\etamax$.
Beyond this learning rate, performance drops sharply.

The fixed compute stopping condition, while of practical interest, biases the results in favor of large learning rates.
Indeed, in the limit of small learning rate, training for a fixed number of steps will keep the model close to initialization.
To control for this, in Figure~\ref{fig:cnn_phys},\ref{fig:wrnL2_phys} models were trained for the same amount of physical time $t\cdot\eta$. For the CNN of figure \ref{fig:cnn_phys}, decaying the learning rate does not have a significant effect on performance and we observe that performance is flat up to $\etamax$, and there is no correlation between our measure of curvature and generalization performance. 
Figure~\ref{fig:wrnL2_phys} shows the analogous experiment for WRN. When decaying the learning rate toward the end of training to control for SGD noise, we find that optimal performance is achieved above $\etacrit$.
In all these cases, $\etamax$ is a good predictor of the maximal learning rate, despite significant differences in the architectures.
Notice that by tuning the learning rate to the catapult phase, we are able to achieve performance using MSE loss, and without momentum, that is competitive with the best reported results for this model \citep{WRN}.

In SM \ref{section:CIFAR100}, we present additional results for WRN on CIFAR-100, with similar conclusions as those for WRN on CIFAR-10.

\section{Additional properties of the model}
\label{sec large width}
So far we have focused on the generalization performance and curvature of the large learning rate phase. 
Here we investigate additional predictions made by our model. 

\subsection{Restoration of linear dynamics}
One striking prediction of the model is that after a period of excursion, the logit differences settle back to $\cO(1)$ values, the NTK stops changing, and evolution is again well approximated by a linear model with constant kernel at large width.

\begin{figure}[h!]
\centering
\ifarxiv
\includegraphics[width=0.5\textwidth]
    {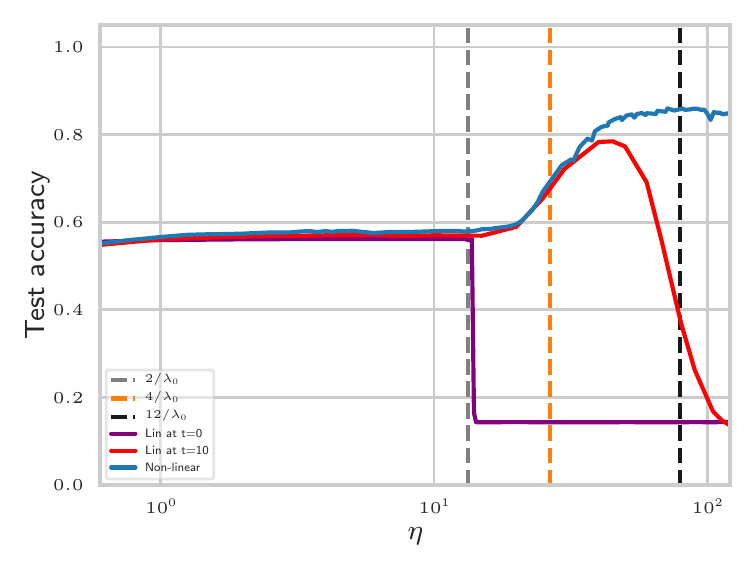}
\else
  \includegraphics
    {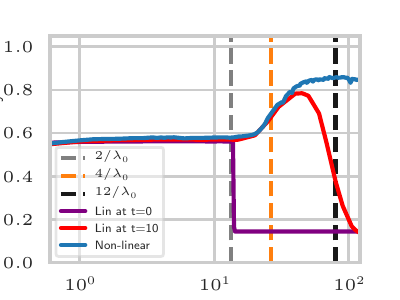}
\fi
\vspace{-4mm}
\caption{
Evidence for linear dynamics after the catapult effect is over. Here we show the same model as in Figure~\ref{fig:fcperfgap} with the addition of models linearized at step $0$ and another linearized at step $10$. We observe that the model linearized after $10$ steps tracks the non-linear performance in the catapult phase up to $\eta \approx 4/\lambda_0$.}
\label{fig:fcperfgaplin}

\end{figure}

We speculate that the return to linearity and constancy of the kernel may hold asymptotically in width for more general models for a range of learning rates above $\etacrit$. 
We test this by evolving the model for order $\log(n)$ steps until the catapult effect is over, linearizing the model, and comparing the evolution of the two models beyond this point.
Figure~\ref{fig:fcperfgaplin} shows an example of this. At fixed width, the accuracy of the linear and non-linear networks match for a range of learning rates above the transition up to $4/\lambda_0$. We present additional evidence for this asymptotic linearization behavior in the Supplement.

\subsection{Non-perturbative phase transition}
The large width analysis of the small learning rate phase has been the subject of much work. In this phase, at infinite width, the network map evolves as a linear random features model, $f^{(0)}_{t+1}=f_{t}^{(0)}-\Theta f_{t}^{(0)}$, where $f^{(0)}$ is the function of the linearized model. At large but finite width, corrections to this linear evolution can be systematically incorporated via a perturbative expansion (Taylor expansion) around infinite width \cite{feynman-diagrams, nth}.
\begin{align}\label{eq:pert}
    f_{t}=f^{(0)}_{t}+\frac{1}{n}f^{(1)}_{t}+\cdots\,.
\end{align}
The evolution equations \eqref{eq:fupdate} and \eqref{eq:Thetaupdate} of the solvable model are an example of this. At large width and in the small learning rate phase, the $O(n^{-1})$ terms are suppressed for all times. In contrast, the leading order dynamics of $f^{(0)}_t$ diverge when $\eta > \eta_{\rm crit}$, and so the true evolution cannot be described by the linear model. Indeed, the logits grow to $\mathcal{O}(n^{1/2})$ and thus all terms in \eqref{eq:fupdate} and \eqref{eq:Thetaupdate} are of the same order. Similarly, the growth observed empirically in the catapult phase for more general models cannot be described by truncating the series \eqref{eq:pert} at any order, because the terms all become comparable. 

\section{Discussion}

In this work we took a step toward understanding the role of large learning rates in deep learning.
We presented a dynamical mechanism that allows deep networks to be trained at larger learning rates than those accessible to their linear counterparts.
For MSE loss, linear model training diverges when the learning rate is above the critical value $\etacrit = 2/\lambda_0$, where $\lambda_0$ is the curvature at initialization.
We showed that deep networks can train for larger learning rates by navigating to an area of the landscape that has sufficiently low curvature.
Perhaps counterintuitively, training in this regime involves an initial period during which the loss increases before converging to its final, small value.
We call this the {\em catapult effect}.

\subsection{A tractable model illustrating catapult dynamics}

These observations are made concrete in our theoretical model, where we fully analyze the gradient descent dynamics as a function of the learning rate.
The analysis involves a modified large width limit, in which both the width and training time are taken to be large.
Sweeping the learning rate from small to large, and working in the limit, we find sharp transitions from a {\em lazy phase} where linearized model training is stable, to a {\em catapult phase} in which only the full model converges, and finally to a {\em divergent phase} in which training is unstable.
These transitions have the hallmarks of phase transitions that commonly appear in physical systems such as ferromagnets or water, as one changes parameters such as temperature. 
In particular, these transitions are non-perturbative: a Taylor series expansion of the linearized model that takes into account finite width corrections is not sufficient to describe the behavior beyond the critical learning rate.

We derive the learning rates at which these transitions occur as a function of the curvature at initialization.
We then treat these theoretical results as predictions, to be tested beyond the regime where they are guaranteed to hold, and find good quantitative agreement with empirical results across a variety of realistic deep learning settings.

We find it striking that a relatively simple theoretical model can correctly predict the behavior of realistic deep learning models.
In particular, we conjecture that the maximum learning rate is typically a simple function of the curvature at initialization, with a single parameter $\carch$ that seems to depend only on the non-linearity.
For ReLU networks, we conjecture that the maximum learning rate is approximately $12/\lambda_0$, which we confirm in many cases.

\subsection{Reducing misalignment of activations and gradients}

The catapult dynamics for the simplified model in Section \ref{sec warmup} reduce curvature by shrinking the component of the first layer weights $u$ which is orthogonal to the second layer weights $v$, and shrinking the component of the second layer weights $v$ which is orthogonal to the first layer weights $u$. 
We can rewrite the simplified model in terms of a hidden layer $h = ux$, where $f(x) = n^{-1/2} v^\top h$. The gradient with respect to this hidden layer is $\pd{L}{h} = n^{-1/2} f(x) v$. 
These hidden layer gradients $\pd{L}{h}$ thus point in the same direction as $v$, while
the hidden activations $h$ point in the same direction as $u$. 
An alternative interpretation of the catapult dynamics is then that they reduce the components of $h$ and $\pd{L}{h}$ which are orthogonal to each other. 
The catapult dynamics thus serve, in this simplified model, to reduce the misalignment between feedforward activations $h$, and backpropagated gradients $\pd{L}{h}$. 
We hypothesize that this reduction of misalignment between activations and gradients may be a feature of large learning rates and catapult dynamics in deep, as well as shallow, networks. We further hypothesize that it may play a directly beneficial role in generalization, for instance by making the model output less sensitive to orthogonal, out-of-distribution, perturbations of activations.

\subsection{Catapult dynamics often improve generalization}

Our results shed light on the regularizing effect of training at large learning rates. 
The effect presented here is independent of the regularizing effect of stochastic gradient noise, which has been studied extensively.
Building on previous works, we noted the observed correlation between flatness and generalization performance.
Based on these observations, we expect the optimal performance to often occur for learning rates larger than $\etacrit$, where the linearized model is unstable.
Observing this effect required controlling for several confounding factors that affect the comparison of performance between different learning rates.
Under a fair comparison, and also for a fixed compute budget, we find that this expectation holds in practice.

\subsection{Beyond infinite linear models}

One outcome of our work is to 
address the performance gap between ordinary neural networks, and linear models inspired by the theory of wide networks.
Optimal performance is often obtained at large learning rates which are inaccessible to linearized models.
In such cases, we expect the performance gap to persist even at arbitrarily large widths.
We hope our work can further improve the understanding of deep learning methods.

\subsection{Other open questions}
There are several remaining open questions.
While the model predicts a maximum learning rate of $4/\lambda_0$, for models with ReLU activations we find that the maximum learning rate is consistently higher.
This may be due to a separate dynamical curvature-reduction mechanism that relies on ReLU.
In addition, we do not explore the degree to which our results extend to softmax classification. 
While we expect qualitatively similar behavior there, the non-constant Hessian of the softmax cross entropy makes controlled experiments more challenging.
Similarly, behavior for other optimizers such as SGD with momentum may differ.
For example, the maximum learning rate when training a linear model is larger for gradient descent with momentum than for vanilla gradient descent, and therefore the transition to the catapult phase (if it exists) will occur at a higher learning rate.
We leave these questions to future work.

\section*{Acknowledgements}

The authors would like to thank 
Kyle Aitken,
Dar Gilboa, 
Justin Gilmer, 
Boris Hanin,
Tengyu Ma, 
Andrea Montanari,
and Behnam Neyshabur
for useful discussions. We would also like to thank Jaehoon Lee for early discussions about empirical properties of the lazy phase. 



\bibliographystyle{icml2020}

\newpage
\setcounter{equation}{0}
\setcounter{figure}{0}
\setcounter{table}{0}
\setcounter{page}{1}
\setcounter{section}{0}
\renewcommand{\theequation}{S\arabic{equation}}
\renewcommand{\thefigure}{S\arabic{figure}}
\renewcommand{\thetable}{S\arabic{table}}

\section*{Supplementary materials}
\appendix
\section{Experimental details}
\label{sec:expdetails}
We are using JAX \citep{jax2018github} and the Neural Tangents Library for our experiments \citep{neuraltangents2020}. 

All the models have been trained with Mean Squared Error normalized as ${\cal L}(\lbrace x,y \rbrace_B) =\frac{1}{2 k |B|}\sum_{(x,y) \in B,i} (f^i(x)-y^i)^2 $, where $k$ is the number of classes and $y^i$ are one-targets. 

In a similar way, we have normalized the NTK as $\Theta_{i j}(x,x')=\frac{1}{k |B|} \sum_{\alpha} \partial_{\alpha} f^{i}(x) \partial_{\alpha} f^{j}(x')$ so that the eigenvalues of the NTK are the same as the non-zero eigenvalues of the Fisher information: $\frac{1}{ k |B|}\sum_{x \in B,i} \partial_{\alpha} f^i(x) \partial_{\beta} f^{i}(x)$.

In our experiments we measure the top eigenvalue of the NTK using Lanczos' algorithm. We construct the NTK on a small batch of data, typically several hundred samples, compute the top eigenvalue, and then average over batches. In this work, we do not focus on precision aspects such as fluctuations in the top eigenvalue across batches.

All experiments that compare different learning rates use the same seed for the weights at initialization and we consider only one such initialization (unless otherwise stated) although we have not seen much variance in the phenomena described. We let $\sigma_w, \sigma_b$ denote the constant (width-independent) coefficient of the standard deviation of the weight and bias initializations, respectively.

Here we describe experimental settings specific to a figure.

\textbf{Figure \ref{fig:FCa},\ref{fig:FCc},\ref{fig:FCe}.} Fully connected, three hidden layers $w=2048$, ReLU  non-linearity trained using SGD (no momentum) on MNIST. Batch size$=512$, using NTK normalization, $\sigma_w = \sqrt{2}, \sigma_b = 0$.

\textbf{Figures \ref{fig:FCb},\ref{fig:FCd},\ref{fig:FCf}.} Wide ResNet 28-18 trained on CIFAR10 with SGD (no momentum). Batch size of $128$, LeCun initialization with $\sigma_w = \sqrt{2}, \sigma_b = 0$, $L_2=0$.

\textbf{Figures \ref{fig:fcperfgap},\ref{fig:fcperfgaplin}} Fully connected network with one hidden layer and ReLU non-linearity trained on 512 samples of MNIST with SGD (no momentum). Batch size of $512$, NTK initialization with $\sigma_w = \sqrt{2}, \sigma_b = 0$. 

\textbf{Figures \ref{fig:cnn_fixed},\ref{fig:cnn_phys}.} The convolutional network has the following architecture: 
$\text{Conv}_{1}(320) \rightarrow \text{ReLU} \rightarrow \text{Conv}_{2}(320) \rightarrow \text{ReLU} \rightarrow \text{MaxPool((2,2), 'VALID')} \rightarrow \text{Conv}_{1}(320) \rightarrow \text{ReLU} \rightarrow \text{Conv}_{2}(128) \rightarrow \text{MaxPool((2,2), 'VALID')} \rightarrow \text{Flatten()} \rightarrow \text{Dense}(256) \rightarrow \text{ReLU} \rightarrow \text{Dense}(10)$. $\text{Dense}(n)$ denotes a fully-connected layer with output dimension $n$. $\text{Conv}_{1}(n), \text{Conv}_{2}(n)$ denote convolutional layers with 'SAME' or 'VALID' padding and $n$ filters, respectively; all convolutional layers use $(3,3)$ filters.  MaxPool((2,2), 'VALID') performs max pooling with 'VALID' padding and a (2,2) window size. LeCun initialization is used, with the standard deviation of the weights and biases drawn as $\sigma_{w} = \sqrt{2}$, $\sigma_{b} = 0.05$, respectively. Trained on CIFAR-10 with SGD, batch size of 256 and L2 regularization = 0.001.

\textbf{Figures \ref{fig:marquee}, \ref{fig:wrnL2_fixed},\ref{fig:wrnL2_phys}.} Wide ResNet on CIFAR10 using SGD (no momentum). Training on v3-8 TPUs with a total batch size of $1024$ (and per device batch size of $128$). They all use $L_2$ regularization$=0.0005$, LeCun initialization with $\sigma_w = 1, \sigma_b = 0$. There is also data augmentation: we use flip, crop and mixup.  With softmax classification, these models can get test accuracy of $0.965$ if one uses cosine decay, so we don't observe a big performance decay due to using MSE. Furthermore, we are using  JAX's implementation of Batch Norm which doesn't keep track of training batch statistics for test mode evaluation. We have not hyperparameter tuned for learning rates nor $L_2$ regularization parameter. 

\textbf{Figures \ref{fig:wrnL2cifar100},\ref{fig:CIFAR100physicalperf}.} Wide ResNet on CIFAR100 using SGD (no momentum). Same setting as figure \ref{fig:wrnL2_fixed}, \ref{fig:wrnL2_phys} except for the different dataset, different L2 regularization $=0.000025$ and label smoothing (we have subtracted $0.01$ from the target one-hot labels).

\textbf{Figure \ref{fig:relu}.}  Two hidden layer, ReLU network for one data point $x=1,y=1$.

\textbf{Figure \ref{fig:tanh}.} Fully connected network with two hidden layers and tanh non-linearity trained on MNIST with SGD (no momentum). Batch size of $512$, LeCun initialization with $\sigma_w = 1, \sigma_b = 0$. 

\textbf{Figure \ref{fig:momdiffact}.} Two-hidden layer fully connected network trained on MNIST with batch size $512$, NTK normalization with $\sigma_w = \sqrt{2}, \sigma_b = 0$. Trained using both momenta $\gamma=0.9$ and vanilla SGD for three different non-linearities: tanh, ReLU and identity (no non-linearity). The learning rate for each non-linearity was chosen to correspond to $\eta=\frac{1}{\lambda_0}$.

\textbf{Rest of SM figures.} Small modifications of experiments in previous figures, specified in captions.

\begin{figure*}[t]
  \centering  
  \begin{overpic}[width=0.75\textwidth]{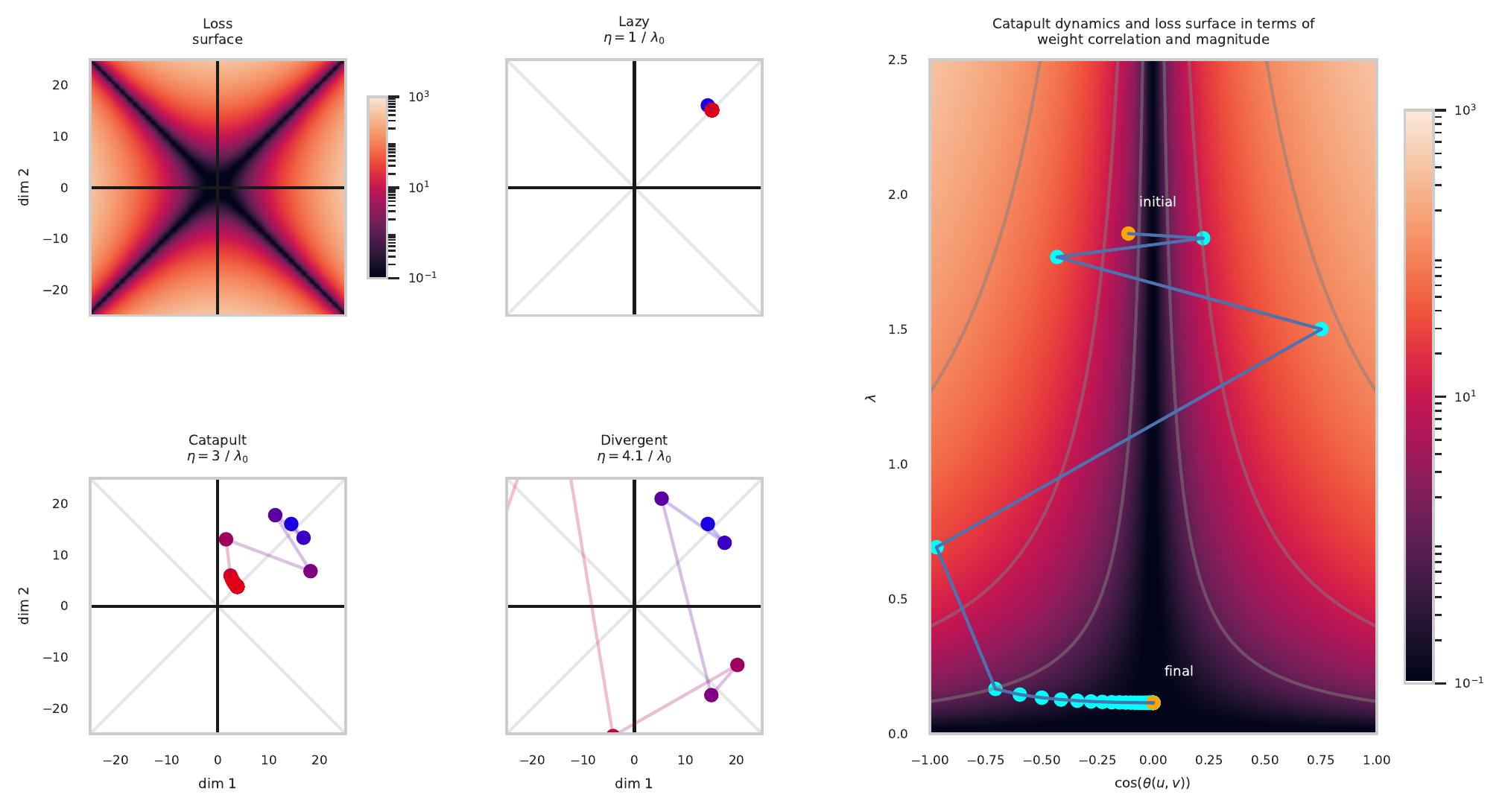}
 \put (0,30) {\small(a)}
 \put (30,30) {\small(b)}
 \put (0,0) {\small(c)}
 \put (30,0) {\small(d)}
 \put (55,0) {\small(e)}
\end{overpic}
  \caption{
    Visualization of training dynamics in all three phases. 
    In the \textbf{lazy phase}, the network is approximately linear in its parameters, and converges exponentially to a global minimum. 
    In the \textbf{catapult phase}, the loss initially grows, while the weight norm and curvature decrease. 
    Once the curvature is low enough, optimization converges. 
    In the \textbf{divergent phase}, both the loss and parameter magnitudes diverge. 
    {\em (a)-(d)} Loss surface and training dynamics
    visualized in
    a 2d linear subspace. 
    The network has a single hidden layer with width $n=500$, linear activations, and is trained with MSE loss on a single 1D sample $x=1$ with label $y=0$.
    The parameter subspace is defined by $u = \operatorname{[dim 1]} r + \operatorname{[dim 2]} s, v = \operatorname{[dim 1]} r - \operatorname{[dim 2]} s$, where $r$ and $s$ are orthonormal vectors, $u,v \in \bR^n$ are the weight vectors, and $\operatorname{[dim 1]}, \operatorname{[dim 2]}$ are the coordinates in the subspace. If initialized in this 2d subspace, $u_t$ and $v_t$ remain in the subspace throughout training, and so training dynamics can be fully visualized with a two dimensional plot.
    {\em (e)}
    Visualization of the loss surface and training dynamics 
    in terms of a nonlinear reparameterization, providing interpretable properties: {\em x-axis} correlation between weight vectors, {\em y-axis} curvature $\lambda$. 
    The trajectory shown is identical to that in (c), and in Figure \ref{fig:marquee}.    
  }
  \label{fig:viz}
\end{figure*}

\section{Experimental results: Late time performance}
\subsection{CIFAR-100 performance}
\label{section:CIFAR100}
\begin{figure}[ht!]
\centering
\subfloat[]{
\includegraphics{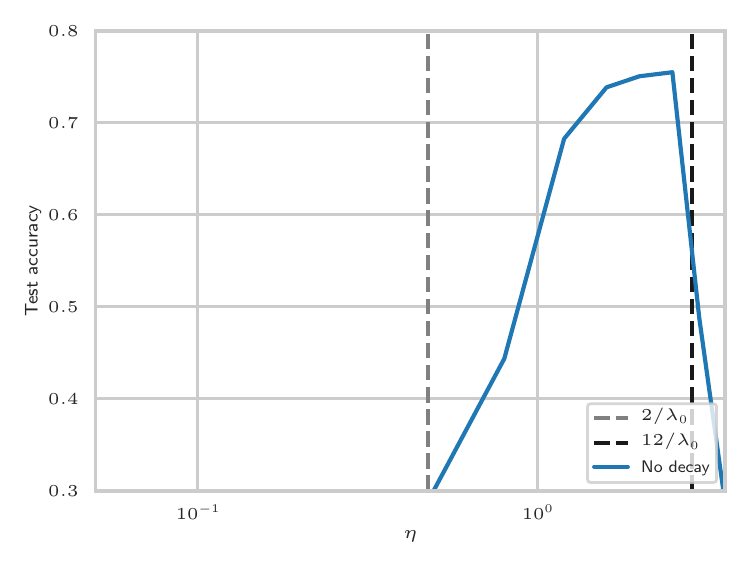}
\label{fig:wrnL2cifar100a}

}
\subfloat[]{
  \includegraphics{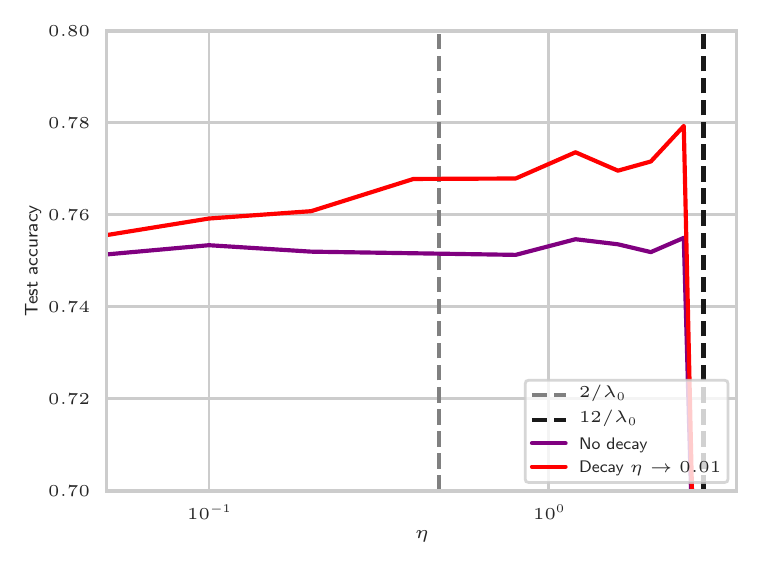}
  \label{fig:wrnL2cifar100b}

}

\caption{Test accuracy vs learning rate for WRN28-10 and CIFAR100 with vanilla SGD, $L_2$ regularization, data augmentation, label smoothing and batch size 1024. The critical learning rate is $\etacrit\approx 0.4$. (a) Evolved for 38400 steps. (b) Evolved for 96000$/\eta$ steps with learning rate $\eta$ (blue) and then evolved for 7200 more steps at learning rate $0.01$ (red). }
\label{fig:wrnL2cifar100}
\end{figure}

We can also repeat the performance experiments for CIFAR-100 and the same Wide ResNet 28-10 setup. In this case, using MSE and SGD we require to evolve the system for longer times, which requires a smaller $L_2$ regularization. We didn't tune for it, but found that $2.5 \times 10^{-5}$ works. With only one decay we can get within $3 \%$ of the \citet{WRN} performance that used softmax classification and two learning rate decays. However, evolution for longer time is needed: we found that different learning rates converge at $\approx 2000$ physical epochs. Similar to the main text experiments, we observe that if we decay after evolving for the same amount of physical epochs, larger learning rates do better. See figure \ref{fig:wrnL2cifar100}.

\subsection{Different learning rates converge at the same physical time}

We can also plot the test accuracy versus physical time for different learning rates to show that for vanilla SGD, the performance curves of different learning rates are basically on top of each other if we plot them in physical time, which is why we find that the fair comparison between learning rates should be at the same physical time. 

\begin{figure}[ht!]
\centering
\subfloat[]{
  \includegraphics{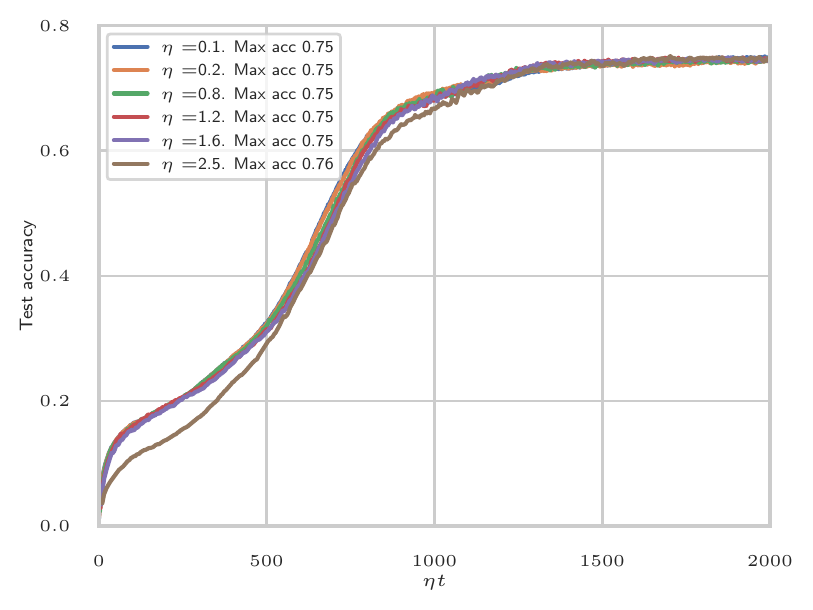}
}

\caption{Test accuracy vs physical time for different learning rates in the WRN CIFAR100 experiment of the previous section \ref{section:CIFAR100} }
\label{fig:CIFAR100physicalperf}
\end{figure}

We have picked a subset of learning rates of the previous WRN28-18 CIFAR100 experiment of SM \ref{section:CIFAR100}. In figure \ref{fig:CIFAR100physicalperf}, we see how even if the curves are slightly different they converge to roughly the same accuracy. The only curve which is slightly different is $\eta=2.5$ which is a rather high learning rate (close to $\frac{12}{\lambda_0}$). 

\subsection{Comparison of learning rates for different $L_2$ regularization for WRN28-10 on CIFAR10}
\label{comparinglr}
Even if in the main section we have considered a model with fixed $L_2$ regularization, we can study the effect without $L_2$ or with a different value. In these two examples, we will be considering the same setup as figures \ref{fig:wrnL2_fixed},\ref{fig:wrnL2_phys}.

Without $L_2$ regularization, we see that the larger learning rate does better even in the absence of learning rate decay, although training takes a really long time. In our experience, comparing this setup with state of the art, $L_2=0$ regularization makes the experiment take longer before convergence but does not influence performance much. 
\begin{figure}[ht!]
  \centering
  \includegraphics{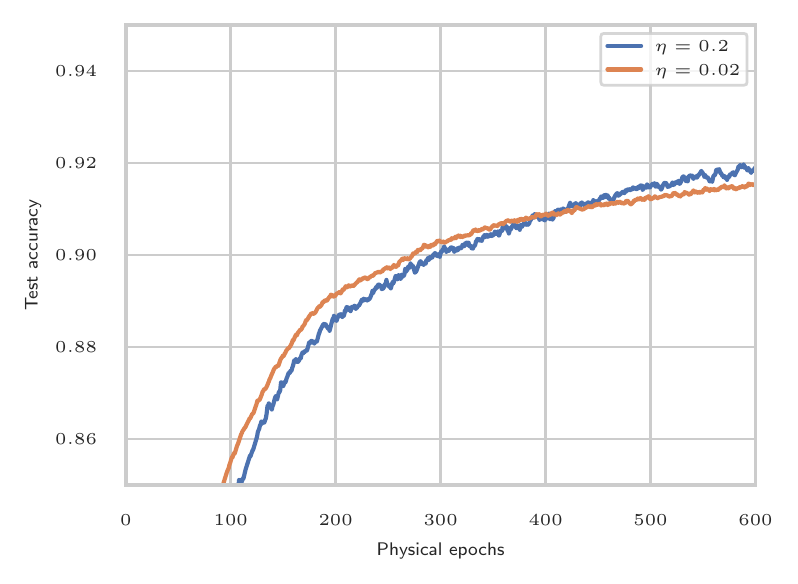}
  \caption{WRN28-10 on CIFAR10 without $L_2$. Same setup as \ref{fig:wrnL2_phys} but evolved for longer times. }
\end{figure}

In the presence of $L_2$ regularization we picked the particular value $L_2=0.0005$
in order to make sure that our conclusion is not dependent on the choice of $L_2$, the only  hyperparameter (other than $\eta$), we have considered a larger $L_2=0.001$. We see that the optimal performance in physical time is also peaked in the catapult phase, although the difference here is smaller. 

\begin{figure}[ht!]
\subfloat[]{
\includegraphics{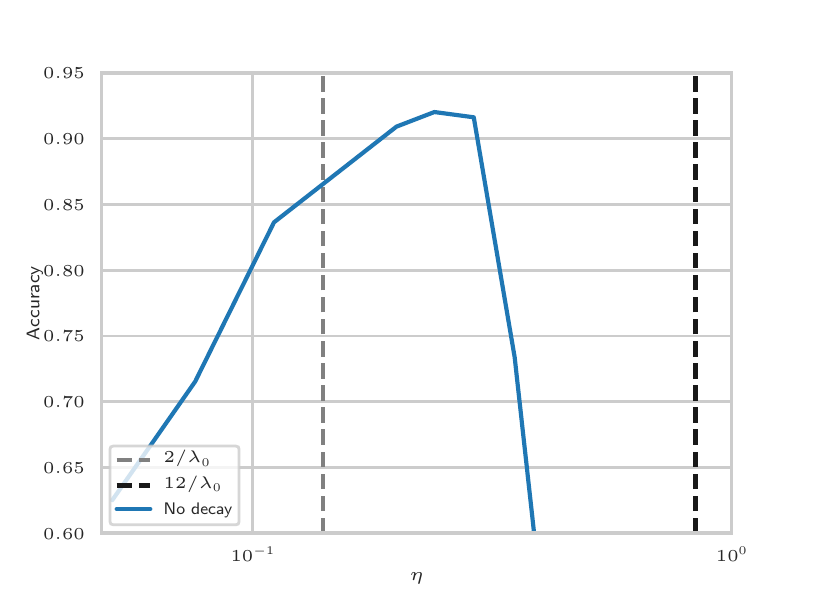}
}
\subfloat[]{
\includegraphics{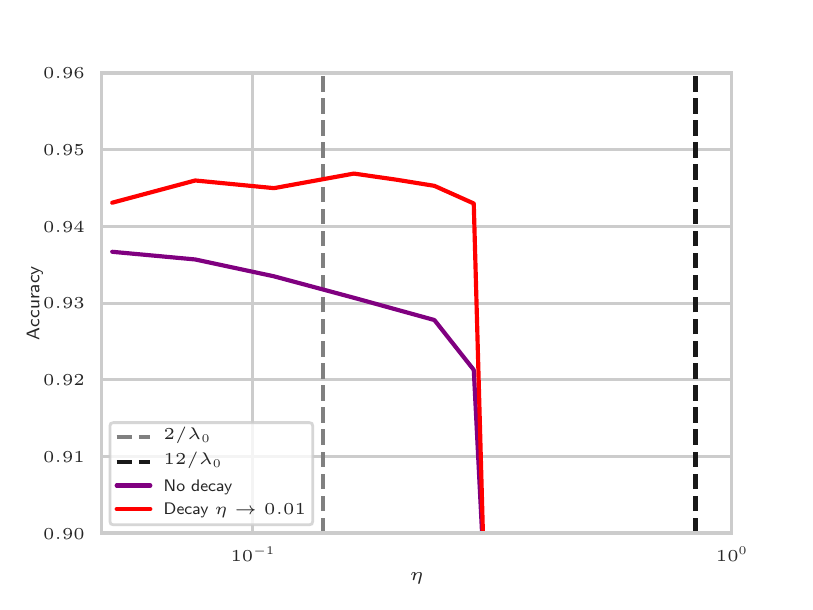}
}

\caption{Test accuracies for a larger $L_2$ CIFAR10 experiment like that of the main section. (a) WRN CIFAR-10 7200 steps as in figure \ref{fig:wrnL2_fixed}. (b) WRN CIFAR10 2400 physical steps and then 4800 more steps at learning rate $0.01$ as in figure \ref{fig:wrnL2_phys}.  }
\label{fig:higherL2}
\end{figure}

\subsection{Training accuracy plots}

\begin{figure}[ht!]
  \centering
  \subfloat[]{
    \includegraphics{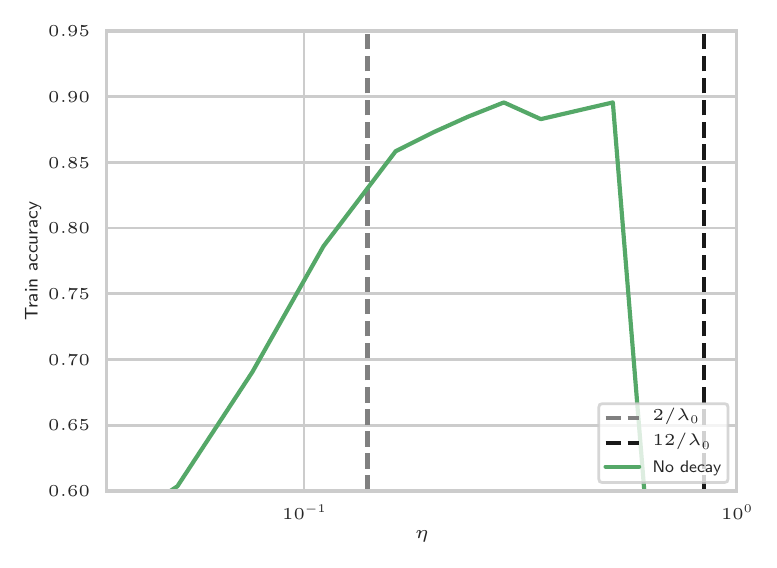}
  }
  \subfloat[]{
    \includegraphics{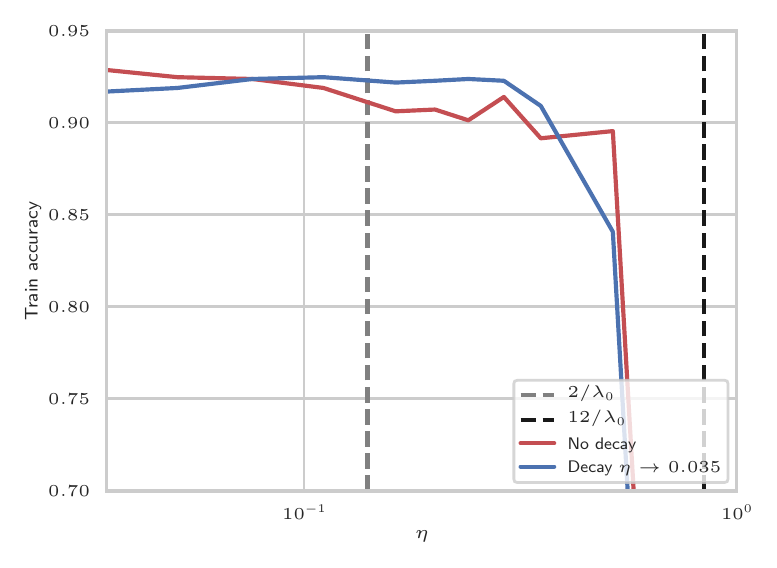}
  }
  \newline
  \subfloat[]{
    \includegraphics{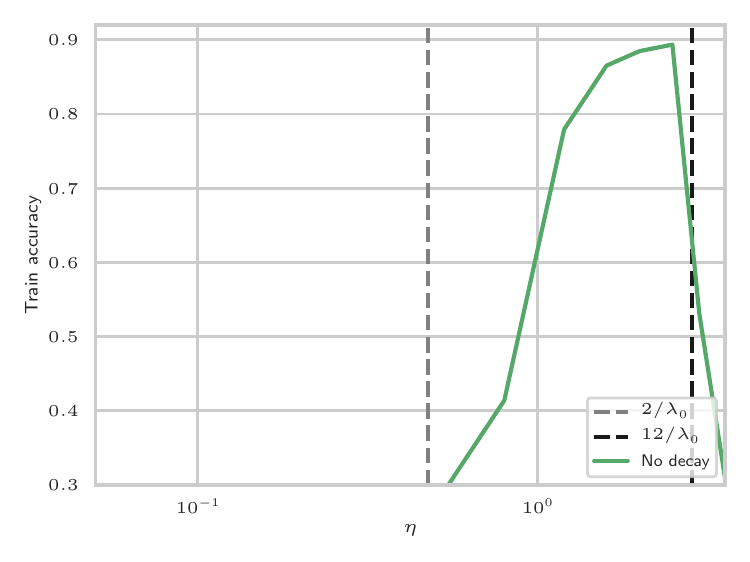}
  }
  \subfloat[]{
    \includegraphics{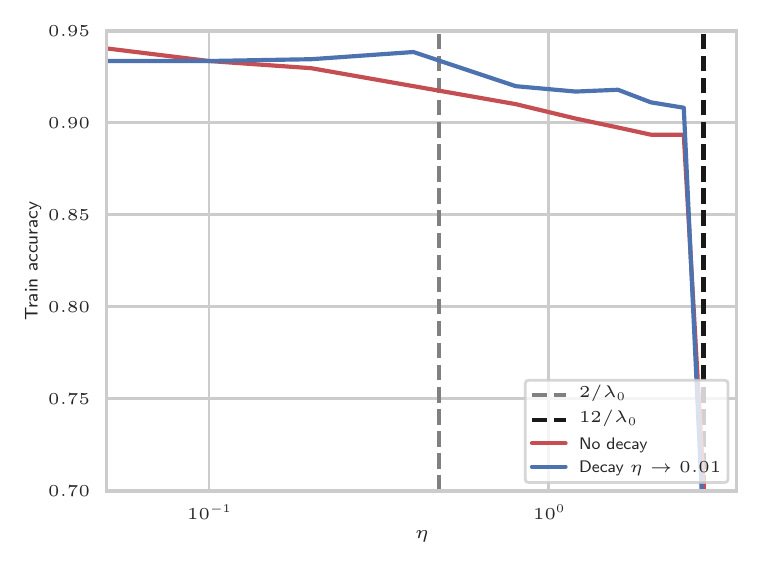}
  }
  \caption{Training accuracies for the performance experiments. Smaller learning rates have higher training accuracy when compared in physical time. However, they still perform worse for a fixed number of steps. (a) WRN CIFAR-10 12000 steps as in figure \ref{fig:wrnL2_fixed}. (b) WRN CIFAR10 3360 physical steps as in figure \ref{fig:wrnL2_phys}. (c) WRN CIFAR100 $38400$ steps as in figure \ref{fig:wrnL2cifar100a}.(d) WRN CIFAR100 $96000$ physical steps as in figure \ref{fig:wrnL2cifar100b}. }
\label{fig:training}
\end{figure}

The training accuracies of the previous experiments are shown in figure \ref{fig:training}. 

\clearpage

\section{Experimental results: Early time dynamics}

\subsection{ReLU activations for the simple model}
\label{sec:relu}
\begin{figure}[ht!]
\centering
\subfloat[]{
  \includegraphics{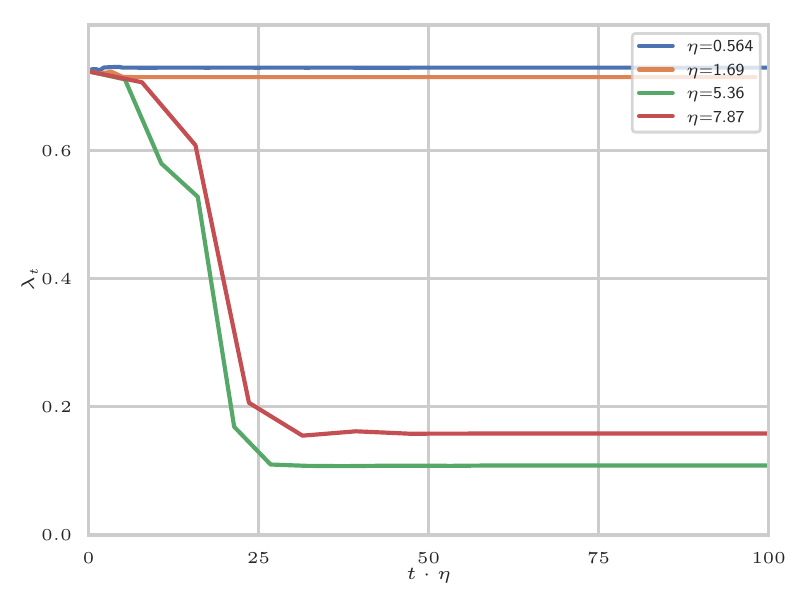}
}
\subfloat[]{
  \includegraphics{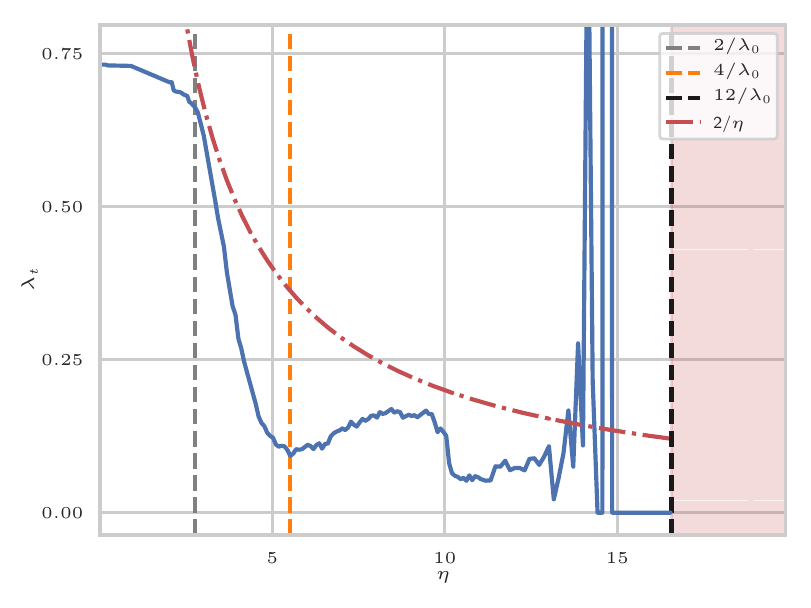}
}

\caption{ Simple model ReLU non-linearity ($\eta_{\rm crit}=2.54$). (b) is evaluated at physical time $100$.   }
\label{fig:relu}
\end{figure}

In the main text we have been using ReLU non-linearities. Compared with the simple model with no non-linearities, ReLU networks have a broader trainability regime after $\eta = \frac{4}{\lambda_0}$. It looks like these networks generically well train until $\eta=\frac{12}{\lambda_0}$. This is a generic feature of deep ReLU networks and can be already observed for the model of section $2$ with a target $y=1$, two hidden layers and a ReLU non-linearity: $f=u.ReLU(w. ReLU(v) )$, as shown in figure \ref{fig:relu}). In this single sample context for $\eta \ge \frac{12}{\lambda}$, the loss doesn't diverge but the neurons die and end up giving the trivial $f=0$ function. For  deep networks with more than one hidden layer and multiple samples, as discussed in the main text, we observe that the loss diverges after $\sim \frac{12}{\lambda}$.

\subsection{Momenta}
\begin{figure}[ht!]
\centering
\subfloat[]{
  \includegraphics[valign=t]{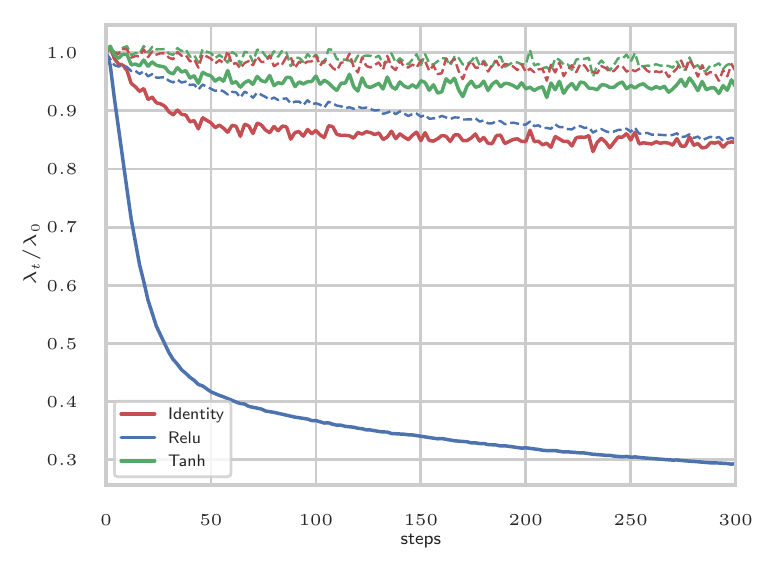}
  \label{fig:momdiffact}
} 
\subfloat[]{
  \includegraphics[width=0.44\textwidth,valign=t]{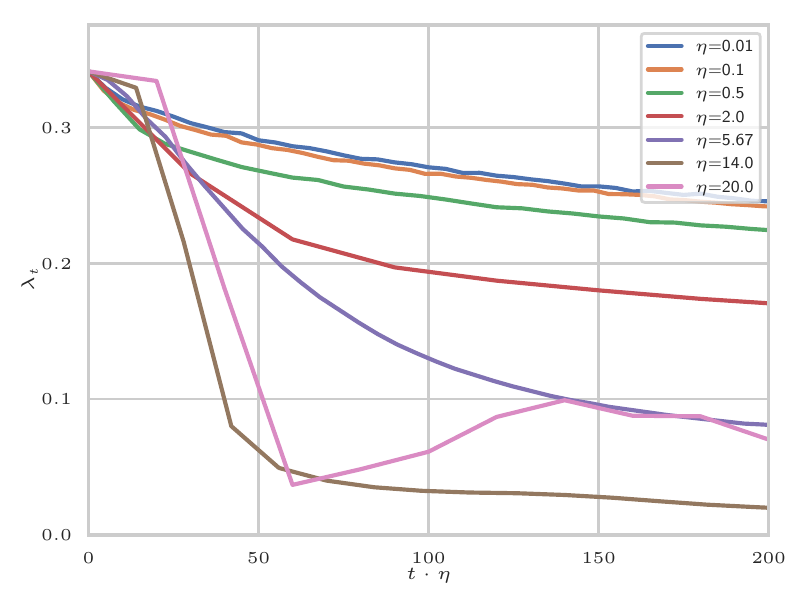}
  \label{fig:momd3relu}
}

\caption{ (a)  Evolution of the normalized curvature $\lambda_t/\lambda_0$ for $d=2$ $w=2048$ FC connected networks evolved with momenta (same networks with SGD with dashed line for reference) evolved for $\eta=\frac{1}{\lambda_0}$. We observe that ReLU networks evolved with momenta doesn't have a constant kernel in the naive `lazy' phase. (b) $\eta_{\rm crit}=6.96, \eta_{\rm crit,eff}=0.69$ Same setup as the FC network of figure \ref{fig:FC} with momenta $\gamma=0.9$: fully connected, three hidden layers $w=2048$, ReLU non-linearity. $\eta_{\rm crit}$ is slightly different due to variations at initialization. }

\end{figure}

The effect of the optimizer also affects these dynamics. If we consider a similar setup with momenta, first we expect that a linear model converges in a broader range $\eta < \frac{2}{\lambda_0} (1+\gamma)$. For smooth non-linearities, we observe that for $\eta < \frac{2}{\lambda_0}$, the $\lambda_t$ is constant. However this is not true for ReLU, see figure \ref{fig:momdiffact}.
In fact, for ReLu networks, we observe that there is a small learning rate, roughly $\eta_{\rm eff,crit}=\frac{\eta_{\rm crit}}{1- \gamma}$, below which the time dynamics of $\lambda_t$ is similar (but non-constant). However, for $\eta > \eta_{\rm eff,crit}$, there are strong time dynamics, we illustrate this in figure \ref{fig:momd3relu} with a 3 hidden layer ReLu network.

\subsection{Effect of $L_2$ regularization to early time dynamics}
\label{section:wrnL2early}
We don't expect $L_2$ regularization to affect the early time dynamics, but because of the strong rearrangement that goes on in the first steps, it could potentially have a non-trivial effect; among other things, the Hessian spectrum necessarily is decaying. We can see how the dynamics that drives the rearrangement is roughly the same, even in the maximum eigenvalue at early times is decreasing slowly.

\begin{figure}[ht!]
\centering

\subfloat[]{
  \includegraphics{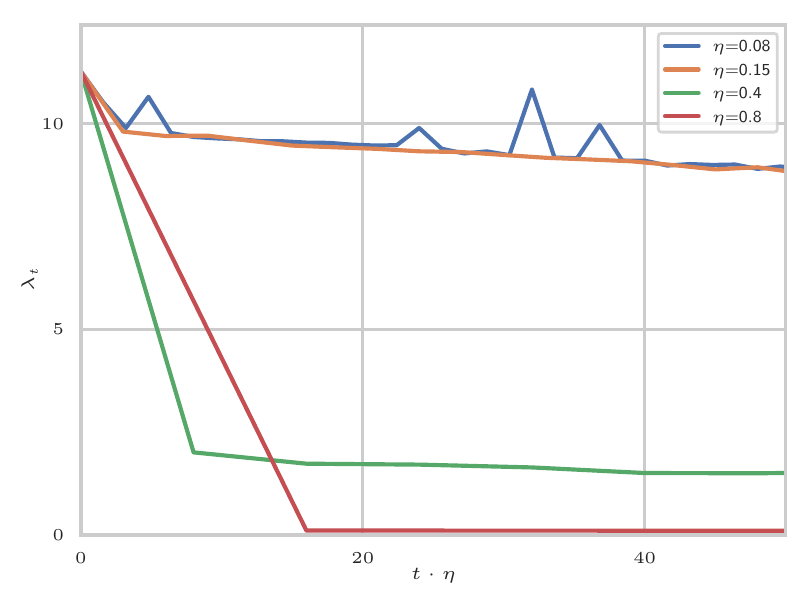}
}
\subfloat[]{
  \includegraphics{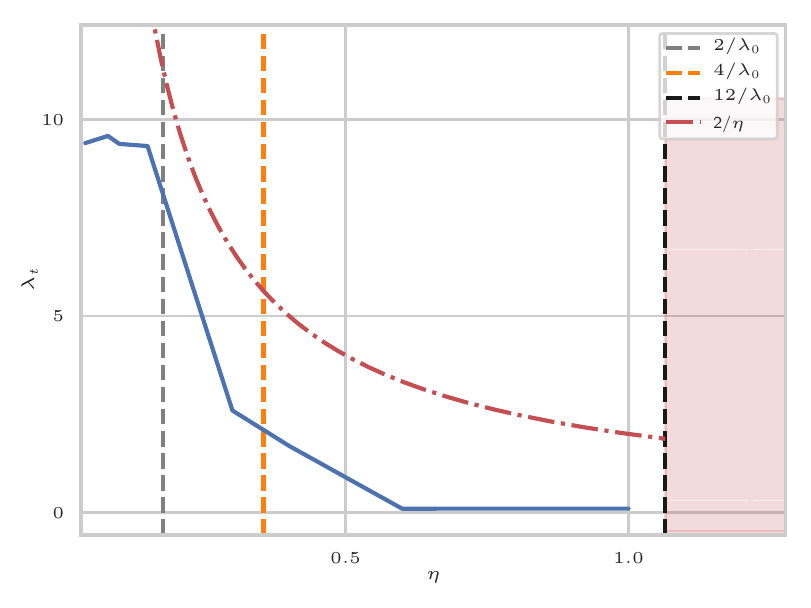}
}
 \caption{Same WRN as figure \ref{fig:FC}d,f with $L_2$ regularization$=0.0005$.  Dynamics in physical steps of the $\lambda_t$ and $\lambda_{t}$ vs $\eta$. $\eta_{\rm crit}=0.18$ a) $\lambda_t$, b) $\lambda_t$ at physical time $25$}
    \label{fig:WRN_L2}
\end{figure}

\subsection{Tanh activations}
\label{sec:tanh}
\begin{figure}[ht!]

  \subfloat[]{
    \includegraphics
    {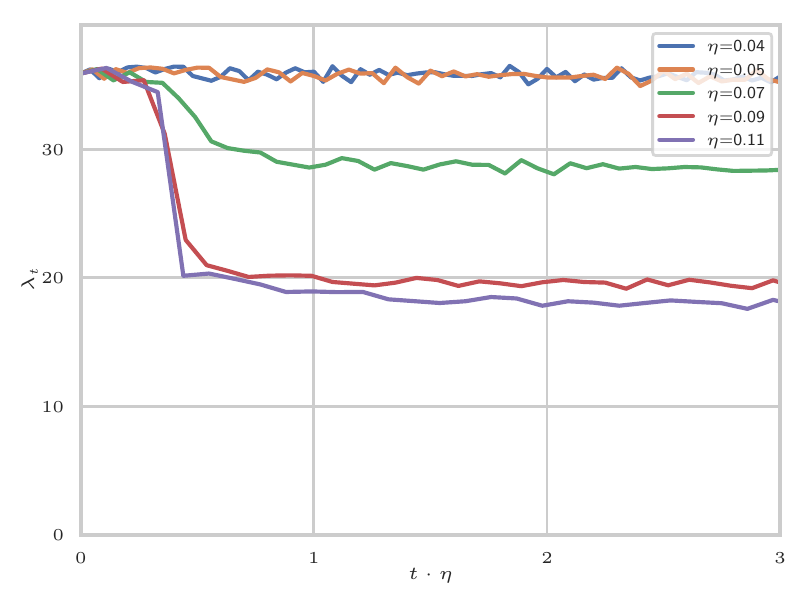}

  }
  \subfloat[]{
    \includegraphics 
    {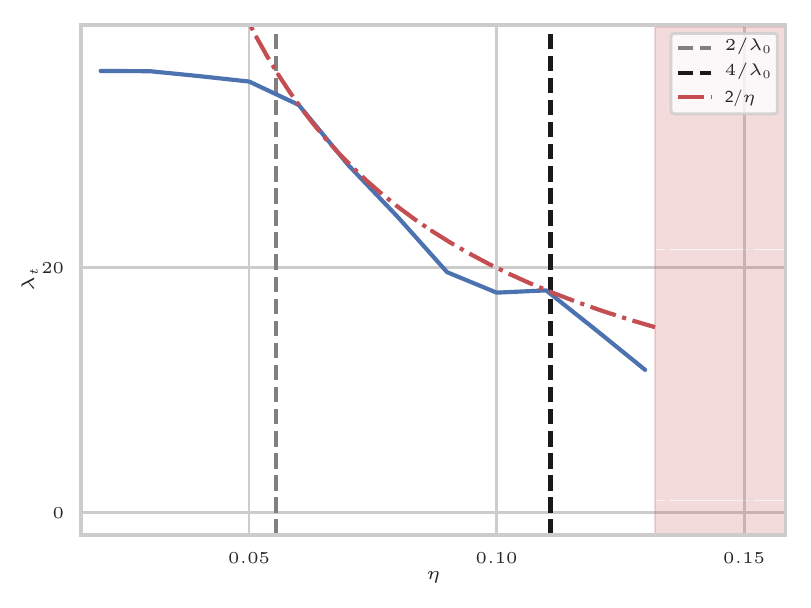}

  }

  \caption{Maximum NTK eigenvalue $\lambda$ at early times for a 2 hidden layer fully connected network with tanh non-linearity trained on MNIST, with $\eta_{\rm crit}=0.06$.
    (a) Early time dynamics of the curvature for learning rates in the linear and catapult phase.
    (b) $\lambda$ measured at $\eta t=3$.
}\label{fig:tanh}
\end{figure}

 We observe that for Tanh activation, $\eta_{\rm max}$ is closer to the simple model expectation $\frac{4}{\lambda_0}$, see figure \ref{fig:tanh}.
\subsection{WRN NTK Normalization}
\begin{figure}[ht!]

\centering

\subfloat[ ]{
  \includegraphics{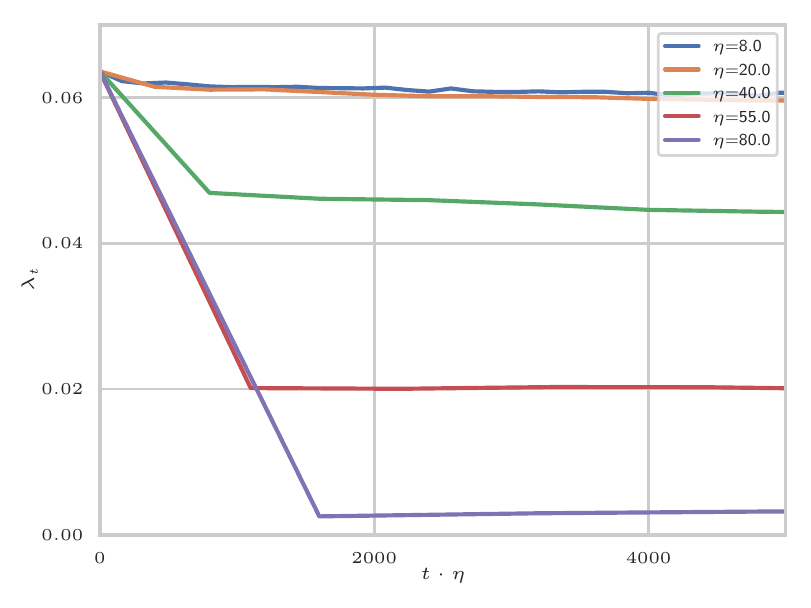}
}
\subfloat[]{
  \includegraphics{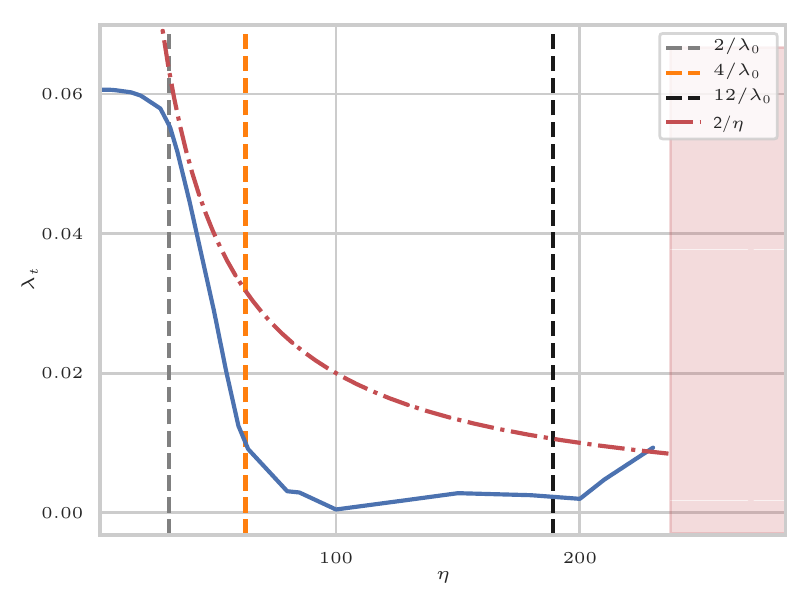}
}
\caption{Same as figures~\ref{fig:FCd},\ref{fig:FCf} but with NTK normalization. a,b) Wide Resnet 28-10. $\eta_{\rm crit}=31.47$,$\lambda$ vs $\eta$ at physical time $4000$}
\label{fig:NTKnorm}
\end{figure}

As illustrated in the text in figures $\ref{fig:FCc},\ref{fig:FCe}$ we also see this behaviour for NTK normalization. For completeness we include the WRN model with NTK normalization.
From the linearized intuition, we expect the phases to also be determined by the quantity $\eta \lambda_{t}$, independently of the normalization. Figure \ref{fig:NTKnorm} has the same setup as in figure~\ref{fig:FC}. 

\clearpage
\section{Theoretical details}

\subsection{Full model analysis}
\label{sec:sm:fullmodel}
Here we provide additional details on the theoretical analysis of the full model in Section~\ref{sec:full_model}.
The gradient descent update equations are
\begin{align}
  u^{t+1}_{ia} &= u_{ia} - \frac{\eta}{\sqrt{n} m} v_i x_{a \alpha} \Df_\alpha \,,\quad
  v^{t+1}_i = v_i - \frac{\eta}{\sqrt{n} m} u_{ia} x_{a \alpha} \Df_\alpha \,.
\end{align}
and 
\begin{equation}
    \Theta_{\alpha \beta}=\frac{1}{n m} (|v|^2 x^T_{\alpha} x_{\beta}+x_{\alpha}^T u^T u x_{\beta} )
\end{equation}
The update equations for the error and kernel evaluated on training set inputs are
\begin{align}
  \Df_\alpha^{t+1} &= (\delta_{\alpha\beta} - \eta \Theta_{\alpha\beta}) \Df_\beta + \frac{\eta^2}{nm} (x_\alpha^T \zeta) (f^T \Df) \,,
  \label{eq:sm:Df}
  \\
  \Theta^{t+1}_{\alpha\beta} &= \Theta_{\alpha\beta} - \frac{\eta}{nm} \left[
     (x_\beta^T \zeta) f_\alpha + (x_\alpha^T \zeta) f_\beta
     + \frac{2}{m} (x_\alpha^T x_\beta) (\Df^T f)
     \right]
     \cr &\quad
           + \frac{\eta^2}{n^2 m} \left[
           |v|^2 (x_\alpha^T \zeta) (x_\beta^T \zeta) + (\zeta^T u^T u \zeta) (x_\alpha^T x_\beta)
           \right] \,.
\end{align}
 Where $\zeta := \sum_\alpha \Df_\alpha x_\alpha / m \in \bR^d$.
We now consider the dynamics of the kernel projected onto the $\Df$ direction, which is given by
\begin{align}
  \Df^T \Theta_{t+1} \Df &=
  \Df^T \Theta \Df +
  \frac{\eta}{n} \zeta^T \zeta \left( \eta \Df^T \Theta \Df - 4 f^T \Df \right) \,.
  \label{eq:sm:dt}
\end{align}
Let us now analyze the phase structure of \eqref{eq:sm:Df} and \eqref{eq:sm:dt}.
For now, we neglect the last term on the right-hand side of \eqref{eq:sm:Df} (at initialization this term is of order $n^{-1}$ and is negligible at large width).
Let $\lmax_0$ be the maximal eigenvalue of the kernel at initialization, and let $\emax \in \bR^m$ be the corresponding eigenvector.
Notice that $\Df$ projected onto the top eigenvector evolves as
\begin{align}
  (\emax)^T {\Df}_{t+1} &= (1 - \eta \lmax) {\emax}^T \Df + \cO(n^{-1}) \,. \label{eq:sm:eDf}
\end{align}

\paragraph{Lazy phase.} 
When $\eta \lmax_0 < 2$, we see that $| {\emax}^T \Df^{t} |$ shrinks during training.
The kernel updates are of order $n^{-1}$, while convergence happens in order $n^0$ steps.
Therefore the kernel does not change by much during training.
This is a special case of the NTK result \citep{NTK-paper}.
Effectively, the model evolves as a linear model in this phase.

\paragraph{Catapult phase.} 
When $2 < \eta \lmax_0 < 4$, $\| \Df \|_2$ grows exponentially fast, and it grows fastest in the $\emax$ direction.
Therefore, the vector $\Df$ becomes aligned with $\emax$ after a number of steps that is of order $n^0$.
Also, $f$ itself grows quickly while the label is constant, and so we find that $f \approx \Df \approx ({\emax}^T \Df) \emax$ after a similar number of steps.
When these approximations hold, notice that $\Df^T \Theta \Df \approx \lambda \cdot \|\Df\|_2^2$.
From equation \eqref{eq:sm:dt} we can then derive an approximate equation for the evolution of the top NTK eigenvalue.
\begin{align}
  \lmax_{t+1} &\approx \lmax + \frac{\eta}{n} \zeta^T \zeta ( \eta \lmax - 4 ) \,.
\end{align}
While $\Df$ grows exponentially fast, so will $\zeta$.
When $\zeta_t$ becomes of order $n^{1/2}$, the updates to the top eigenvalue become of order $n^0$ (and negative), causing $\lambda_t$ to decrease by a non-negligible amount.
This will continue until $\lmax_t < 2/\eta$, at which point $\Df_t$ will start converging to zero.
Eventually, after a number of steps of order $\log(n)$, gradient descent will converge to a global minimum that has a lower curvature than the curvature at initialization.

The justification for dropping the order $n^{-1}$ term in \eqref{eq:sm:eDf} was explained in the warmup model: While this term may affect the details of the dynamics, eventually the maximum kernel eigenvalue must drop below $2/\eta$ for the component ${\emax}^T \Df$ of the error (and therefore for the loss) to converge to zero.

\paragraph{Divergent phase.} 
When $\eta \lmax_0 > 4$, both $\| \Df \|_2^2$ and $\lmax$ will grow, and optimization will diverge.
Therefore, $4/\lmax_0$ is the maximum learning rate for this model.

\section{Model dynamics close to the critical learning rate}

Here we consider the gradient descent dynamics of the model analyzed in Section~\ref{sec:model}, for learning rates $\eta$ that are close to the critical point $\etacrit=2/\lambda_0$.
The analysis reveals that the gradient descent dynamics of the model are qualitatively different above and below this point.
For example, the loss decreases monotonically during training when $\eta < \etacrit$, but not when $\eta > \etacrit$.
In this section we show that the transition from small to large learning rate becomes sharp once we take the modified large width limit, in the following sense: certain functions of the learning rate become non-analytic at $\etacrit$ in the limit.
This sharp transition bears close resemblance to phase transitions of the kind found in physical systems, such as the transition between the liquid and gaseous phases of water.
In particular, our case involves a dynamical system, where the dynamics are governed by the gradient descent equations.
These dynamics undergo a phase transition as a function of the learning rate --- an external parameter.
We point to the logistic map \cite{may1976simple} as a well-known example of a dynamical system that undergoes phase transitions as a function of an external parameter.

\subsection{Non-perturbative dynamics}
A phase transition is a drastic change in a system's behavior incurred under a small change in external parameters.
Mathematically, it is a non-analyticity in some property of the system as a function of these parameters.
For example, consider the property $\lambda_*(\eta)$, the curvature of the model at the end of training as a function of the learning rate.
In the modified large width limit, $\lambda_*(\eta)$ is constant for $\eta < \etacrit$, but not for $\eta > \etacrit$.
Therefore, this function is not analytic at $\etacrit$.
Notice that this statement is true in the limit but not necessarily at finite width, where the final curvature may be an analytic function of the learning rate even at $\etacrit$.
It is well known in physics that phase transitions only occur in a limit where the number of dynamical variables (in this case the number of model parameters) is taken to infinity.
One immediate consequence of the non-analyticity at $\etacrit$ is that the large learning rate phase is inaccessible from the small learning rate phase via a perturbative expansion.
In other words, we cannot describe all properties of the model for some $\eta > \etacrit$ by doing a Taylor expansion around a point $\eta_0 < \etacrit$ and keeping a finite number of terms.

\citet{feynman-diagrams,nth} developed a formalism that allows one to compute finite-width corrections to various properties of deep networks, using a perturbative expansion around the infinite width limit.
We have argued that the usual infinite width approximation to the training dynamics is not valid for learning rates above $\etacrit$, and that a full analysis must account for large finite-width effects.
One may have hoped that including the perturbative finite-width corrections discussed in \citet{feynman-diagrams,nth} would allow us to regain analytic control over the dynamics.
The results presented here suggest that this is not the case:
For $\eta > \etacrit$, we expect that the perturbative expansion will not provide a good approximation to the gradient descent dynamics at any finite order in inverse width.

\subsection{Critical exponents}
When the external parameters are close to a phase transition, one often finds that the dynamical properties of the system obey power law behavior.
The exponents of these power laws (called \emph{critical exponents}) are of interest because they are often found to be universal, in the sense that the same set of exponents is often found to describe the phase transitions of completely different physical systems.

Here we consider $t_*(\eta)$, the number of steps until convergence, as a function of the learning rate.
We will now show that $t_*$ exhibits power-law behavior when $\eta$ is close to $\etacrit$.
For simplicity we consider the warmup model studied in Section~\ref{sec:model}.
First, suppose that we are below the transition, setting $\eta \lambda_0 = 2 - \epsilon$ for some small $\epsilon > 0$.
From the update equation, $f_{t+1} \approx (1 - \eta \lambda_t) f_t \approx - (1 - \epsilon) f_t$ we see that $f_t$ will converge to some fixed small value $f_*$ after time $t_* \approx \epsilon^{-1} \log (f_*^{-1}) \sim \epsilon^{-1}$.
Here we assumed that $\lambda_t$ is constant in $t$, which is true as long as $t_*$ is independent of $n$ (namely we fix $\epsilon$ and then take $n$ large).
Therefore, the convergence time below the transition scales as $t_* \sim (\etacrit - \eta)^{-1}$, and the critical exponent is -1.

\begin{figure}[ht!]
\centering
\subfloat[]{
    \includegraphics{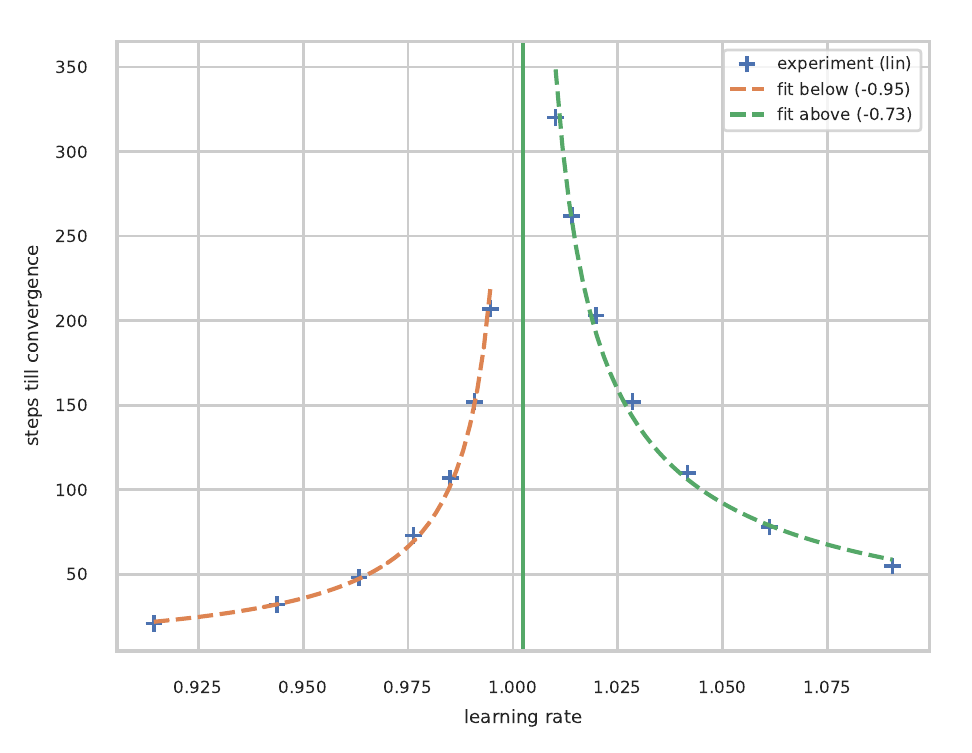}
}
\caption{The convergence time diverges when the learning rate is close to the critical value $\etacrit$, indicated by the solid green line. The measured exponents (shown in parentheses) are close to the predicted value of -1. Experiment involves the warmup model of Section~\ref{sec:model} with width $16,\!000$.
}
\label{fig:critexp}
\end{figure}

Next, suppose that $\eta \lambda_0 = 2 + \epsilon$ with $\epsilon > 0$.
Now the update equation reads $f_{t+1} \approx -(1 + \epsilon) f_t$. 
This approximation holds early during training, when the curvature updates are small.
Initially, $|f_t|$ will grow until it is of order $\sqrt{n}$, at which point the updates to $\lambda_t$ become of order $n^0$.
This will happen in time $\hat{t} \sim \epsilon^{-1} \log \sqrt{n}$.
Following this, the optimizer will converge.
At this point $\eta\lambda_t$ is no longer tuned to be close to the transition, and so the convergence time measured from this point on will not be sensitive to $\epsilon$.
Therefore, for small $\epsilon$ the convergence time will be dominated by the early part of training, namely $t_* \approx \hat{t} \sim \epsilon^{-1}$. 
The critical exponent is again -1.
Figure~\ref{fig:critexp} show an empirical verification of this behavior.

\newpage
\section{Additional evidence for linearization in the catapult phase.}

\begin{figure}[ht!]
  \subfloat[
    ]{
    \includegraphics[valign=t]{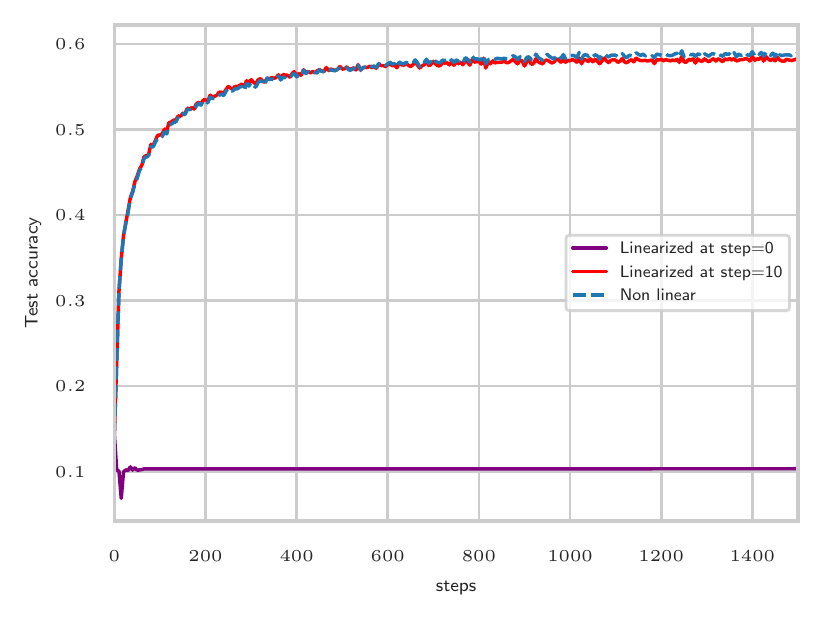}\label{fig:lin_train}
    \vphantom{\includegraphics[valign=t]{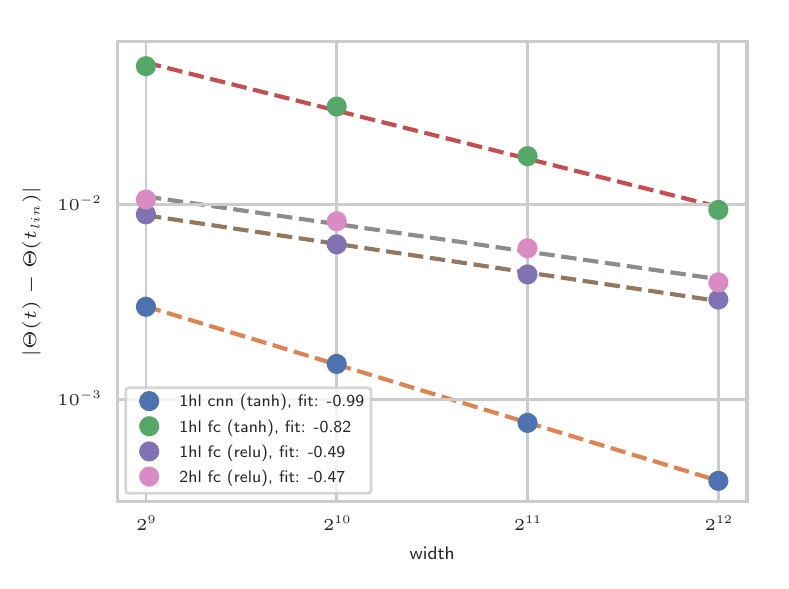}}
  }
    \subfloat[
    ]{
    \includegraphics[valign=t]{figures/ThetaDiff}\label{fig:theta_diff}
  }

\caption{ Evidence for a return of linear dynamics after $t_{\textrm{lin}}$. (a,b) Show the same model as in figure \ref{fig:fcperfgap} with the addition of linearized models at step $0$ and $10$. We observe that the linearized model after 10 steps tracks the non-linear performance in the `catapult' phase up to $\eta \sim \frac{4}{\lambda_0}$ (c) The change in the NTK between $t_{\textrm{lin}}=50$  \textrm{steps} and $t=1000$ steps decreases as the width increases. Here we consider 2-class MNIST with 100 samples per class.}
\label{fig:linearization}

\end{figure}

Here we present some more detailed evidence for the re-emergence of linear dynamics in the catapult phase. Figure~\ref{fig:linearization} show results for models trained on subsets of MNIST with learning rates $\eta>\etacrit$. In figure Figure~\ref{fig:lin_train} we see that for a one-hidden-layer fully connected model trained on 512 MNIST images, the performance of the full non-linear model and model linearized after 10 steps track closely. Models evolve as linear models when the NTK is constant. In Figure~\ref{fig:theta_diff} we give evidence that as networks become wider, the change in the kernel decreases.

\end{document}